\documentclass[journal,final]{IEEEtran}

\usepackage{times}
\usepackage{epsfig}
\usepackage{graphicx}
\usepackage{amsmath}
\usepackage{amssymb}
\usepackage{amsmath}
\usepackage{algorithm}
\usepackage[noend]{algpseudocode}
\usepackage{placeins}
\usepackage{makecell}
\usepackage{multirow}

\usepackage[table,xcdraw]{xcolor}
\usepackage{threeparttable}
\usepackage{subcaption}
\usepackage{colortbl}
\usepackage{multirow}
\usepackage{verbatim}

\usepackage{arydshln}
\usepackage{cite}
\usepackage[export]{adjustbox}

\newcommand{\reviewtxt}[1]{\color{black} #1 \color{black}}
\newcommand{\reviewres}[1]{\color{black} #1 \color{black}}

\ifCLASSINFOpdf
\else
\fi
\usepackage{amsmath,amsfonts,amssymb}
\usepackage{url}

\begin{document}
%
\title{Adversarial Learning and Self-Teaching Techniques for Domain Adaptation in Semantic Segmentation}
%
%
%

\author{Umberto Michieli,~\IEEEmembership{Student Member,~IEEE,}
        ~Matteo~Biasetton, 
        ~Gianluca~Agresti 
        ~and~Pietro~Zanuttigh,~\IEEEmembership{Member,~IEEE}

\thanks{U. Michieli, M. Biasetton, G. Agresti and P. Zanuttigh are with the University of Padova.}
\thanks{Manuscript accepted at IEEE Transactions on Intelligent Vehicles (T-IV)}
\thanks{E-mail: umberto.michieli@dei.unipd.it}}
%
%

\ifCLASSOPTIONpeerreview
\markboth{Journal of IEEE Transactions on XXX,~Vol.~..., No.~..., August~20...}%
{Michieli \MakeLowercase{\textit{et al.}}: Adversarial Learning and Self-Teaching Techniques for Domain Adaptation in Semantic Segmentation}
\fi
%



\maketitle

\begin{abstract}
Deep learning techniques have been widely used in autonomous driving systems for the semantic understanding of urban scenes\reviewtxt{. However, they need a huge amount of labeled data for training, which is difficult and expensive to acquire.} A recently proposed workaround is to train deep networks using synthetic data, but the domain shift between real world and synthetic representations limits the performance. In this work, a novel Unsupervised Domain Adaptation (UDA) strategy is introduced to solve this issue. The proposed learning strategy is driven by three components: a standard supervised learning loss on labeled synthetic data\reviewtxt{;} an adversarial learning module that exploits both labeled synthetic data and unlabeled real data\reviewtxt{; finally,} a self-teaching strategy applied to unlabeled data. The last component exploits a region growing framework guided by the segmentation confidence. Furthermore, we weighted this component on the basis of the class frequencies to enhance the performance on less common classes. Experimental results prove the effectiveness of the proposed strategy in adapting a segmentation network trained on synthetic datasets, like GTA5 and SYNTHIA, to real world datasets like Cityscapes and Mapillary.
\end{abstract}

\ifCLASSOPTIONpeerreview
\fi

%
\IEEEpeerreviewmaketitle

\section{Introduction}
\label{sec:intro}

A key component of any autonomous driving system is the capability of understanding the surrounding environment from visual data. \reviewtxt{Nowadays, this is achieved using semantic segmentation techniques mostly based on deep learning strategies.} 
Deep networks have shown impressive performance on this task\reviewtxt{. However,} they have the key drawback that a huge amount of labeled data is required for their training, especially in case recent highly complex architectures are used.
\reviewtxt{In the autonomous driving scenario,} the pixel-level annotation must be manually provided for a huge amount of frames acquired by  cameras mounted on cars driving around\reviewtxt{. This annotation is expensive and requires a huge amount of work.} 
Some recent papers \cite{Richter2016,ros2016} introduced a workaround for this issue using computer generated data for training the networks.
The realistic rendering models developed by the video game industry can be used to produce a large amount of high quality rendered road scenes \cite{Richter2016}.
\reviewtxt{Despite} the impressive realism of recent video games graphics, there is still a large domain shift between the computer generated data and real world images acquired by video cameras on cars. To be able to really exploit computer generated data in real world applications the domain shift issue needs to be addressed. 
\reviewtxt{Various techniques have been introduced to tackle it: data augmentation techniques try to improve the generalization properties of the available data. Transfer learning approaches allow to adapt the model to the new domain but require some labeled data from the target domain. Finally, Unsupervised Domain Adaptation (UDA) techniques improve the model exploiting unlabeled target data.}
We present an UDA strategy for road driving scenes able to adapt an initial learning performed on synthetic data to the real world scenario. 
The domain adaptation strategy presented in this work is based on adversarial learning and is an extension of our previous work \cite{biasetton2019}: 
here we further improve the self-teaching strategy and we present a more robust experimental evaluation. 
We focus on the training scenario where a large amount of annotated synthetic data is available but there are no labeled real world samples (or just a small amount that can be used for validation purposes but not sufficient for training the deep network).
The proposed method exploits a segmentation network based on the DeepLab v2 framework \cite{chen2018deeplab} that is trained using both labeled and unlabeled data in an adversarial learning framework with multiple components.
The first component that controls the training is a standard  cross-entropy loss exploiting ground truth annotations used to perform a  supervised training  on synthetic data. 
The second is an adversarial learning scheme inspired by previous works on semi-supervised semantic segmentation, i.e., dealing with partially annotated datasets \cite{hung2018, liu2019}. 
 \reviewtxt{We} exploited a fully convolutional discriminator which 
 produces a pixel-level confidence map distinguishing between data produced by the generator (both from real or synthetic data) and the ground truth segmentation maps.
  It allows  to train in an adversarial setting the segmentation network using both  synthetic labeled data and  real world scenes without ground truth information.
Finally, the third term is based on a self-teaching loss. This key component is based on the idea introduced in  \cite{hung2018} that 
the output of the discriminator can also be used  as a measure of the reliability of the network estimations. 
\reviewtxt{This can in turn be exploited to select the reliable regions in a self-teaching framework.}
However, this component has been greatly improved in this work, both with respect to \cite{hung2018} and to 
\cite{biasetton2019}.
First of all, the output of the discriminator has been considered as a weight to be applied to the loss function of the self-teaching component at each location, in place of the hard threshold mask used in previous work \cite{biasetton2019}.
Then, a novel region growing scheme is introduced in order to extend and better represent the shape of reliable regions\reviewtxt{. This is a key difference because the previous approaches \cite{hung2018,biasetton2019} tend to almost always discard edge regions and small objects.}
Finally, since the various classes have different frequencies, we also weighted the loss coming from unlabeled data in proportion to the frequency of the various classes in the synthetic dataset\reviewtxt{. This allows to obtain a better balance of the performance among the different classes. In particular, it avoids dramatic drops in performance on less common classes, as small objects and structures.}

The network has been trained on both synthetic labeled data (using the first and second component) and on unlabeled real world data (using the second and third component) and we were able to obtain accurate results on different real world datasets even without using labeled real  world data.  
In particular, we used the synthetic datasets SYNTHIA and GTA5 for the supervised part and   the real datasets Cityscapes and Mapillary (the latter has been introduced in this journal extension) for the unsupervised components and then tested on the respective validation sets, achieving state-of-the-art results on the unsupervised domain adaptation task.
\section{Related Work}
\label{sec:related}
Many different approaches for semantic segmentation of images have been proposed (see \cite{GarciaGarcia18} for a recent review of the field).
There are many different strategies for this task, but most current state-of-the-art approaches are based on encoder-decoder schemes and on the Fully Convolutional Network  (FCN)  model \cite{long2015}.
Some recent well-known and highly performing  methods are DilatedNet \cite{yu2016}, PSPNet \cite{zhao2017} and DeepLab \cite{chen2018deeplab}. In particular\reviewtxt{,} the latter is the model employed for the generator network in this work. 
All the approaches for generic images can be applied also to road scenes, however since this is a very relevant application \cite{chen2019crdoco, michieli2018} there has been a large effort both in the acquisition of datasets \cite{Cordts2016,neuhold2017mapillary,yu18} and in the development of ad-hoc approaches \cite{zhang2017,chen2018road,kim2018attribute}.

These approaches show impressive performance but they all share the fundamental requirement of having a large amount of labeled data for their training.  They are  typically trained on huge datasets with pixel-wise annotations (e.g.,  Cityscapes \cite{Cordts2016}, CamVid \cite{camvid} or Mapillary \cite{neuhold2017mapillary}), whose acquisition is highly expensive and time-consuming.
Recent research, as the proposed work, focuses on how to deal with this issue both by using only partially labeled data or by adapting the training done on a different set of  data with slightly different statistics to the problem of interest.

 The first family of approaches we consider is semi-supervised methods. They can be divided into methods exploiting weakly annotated data (e.g., with only image-wise labels or only bounding boxes) \cite{pathak2015, souly2017, vezhnevets2010, wei2017, hong2015, dai2015, huang2018, sun2019} or methods for which only part of the data is labeled while the other is completely unlabeled \cite{papandreou2015, sankaranarayanan2018, hung2018, souly2017, liu2019}. 
 The work of \cite{luc2016} has opened the way  to adversarial learning approaches for the semantic segmentation task, while \cite{souly2017}  to their application to semi-supervised learning. The  approaches of \cite{hung2018, liu2019} are also based  on  adversarial learning but exploit  a Fully Convolutional Discriminator (FCD) trying to discriminate between the predicted probability maps and the ground truth segmentation distributions at pixel-level. 
 These works targeted a scenario where only part of the dataset is labeled but unlabeled data comes from the same dataset and shares the same domain data distribution of the labeled ones.

The work of \cite{biasetton2019} 
starts from \cite{hung2018} but instead proposes to tackle a scenario where unlabeled data refers to a different dataset with a  different domain distribution, i.e., it deals with the domain adaptation task.
A common setting for this task is domain adaptation from synthetic data to real world scenes. Indeed, the development of advanced computer graphics techniques enabled the collection of huge synthetic datasets for semantic segmentation. 
Examples of synthetic semantic segmentation datasets for the autonomous driving scenario are the GTA5 \cite{Richter2016} and SYNTHIA \cite{ros2016} datasets, which have been employed in this work. 
However, there is a cross-domain shift that has to be addressed when a neural network trained on synthetic data  processes real-world images, \reviewtxt{since in this case training and test data are not drawn i.i.d. from the same underlying distribution as usually assumed \cite{zhang2017, tommasi2017, torralba2011, gong2012, khosla2012}.} 

\reviewtxt{Data augmentation techniques can be used to improve the generalization capabilities: some works propose to pre-process the synthetic images, i.e., the existing labeled data,} to reduce the inherent discrepancy between real and synthetic domain distributions mainly using generative models based on Generative Adversarial Networks (GANs) \cite{shrivastava2017learning, peng2018synthetic, wang2016generative, zhu2016generative, im2016generating}. \reviewtxt{Thus, these augmented labeled data are used to train the segmentation network to work in a more reliable way on the real domain.}

\reviewtxt{Transfer learning techniques improve the generalization of the trained network, going from the source to the target domain, by using either weak supervision \cite{li2016weakly, inoue2018cross, sun2019} or just a small amount of labeled target data \cite{motiian2017unified, saito2019semi}.}

\reviewtxt{Differently, unsupervised domain adaptation techniques aim at training the networks in a supervised way on synthetic data and in a unsupervised way on real unlabeled data. This family of techniques} has been  widely investigated in classification tasks \cite{ganin2015, ganin2016, long2015learning, tzeng2015simultaneous} but its application to semantic segmentation is less explored.
The first work to deal with cross-domain urban scenes semantic segmentation is \cite{hoffman2016}, where the adaptation is performed by aligning the features from the different during the adversarial training procedure.
A curriculum-style learning approach is proposed in  \cite{zhang2017}, where firstly the easier task of estimating global label distributions is learned and then the segmentation network is trained forcing that the target label distribution is aligned to the previously computed properties.
 Many other works addressed the domain adaptation problem with various techniques, including GANs \cite{sankaranarayanan2018,agresti2019unsupervised}, output space alignment \cite{tsai2018, luo2019}, distillation loss \cite{chen2018road,ICCVW2019}, class-balanced self-training \cite{zou2018}, conservative loss \cite{zhu2018}, geometrical guidance \cite{chen2018geometric}, adaptation networks \cite{zhang2018fully}, entropy minimization \cite{vu2019advent} and cycle consistency \cite{hoffman2018, chen2019crdoco}. This latter technique applies also some kind of data augmentation to the synthetic data to make them more realistic.

Region growing techniques 
have been recently applied to domain adaptation in semantic segmentation \cite{sun2019, huang2018}. In particular in \cite{huang2018} a semantic segmentation network is trained to segment the discriminative regions first and to progressively increase the pixel-level supervision by seeded region growing \cite{adams1994seeded}. In \cite{sun2019} the authors propose a saliency guided weakly supervised segmentation network which utilizes salient information as guidance to help weakly segmentation through a seeded region growing procedure. In \cite{song2018seednet} the region growing problem is represented as a Markov Decision Process.
\section{Architecture of the Proposed Approach}
\label{sec:method}

\begin{figure*}[ht]
\centering
\includegraphics[width=1\textwidth]{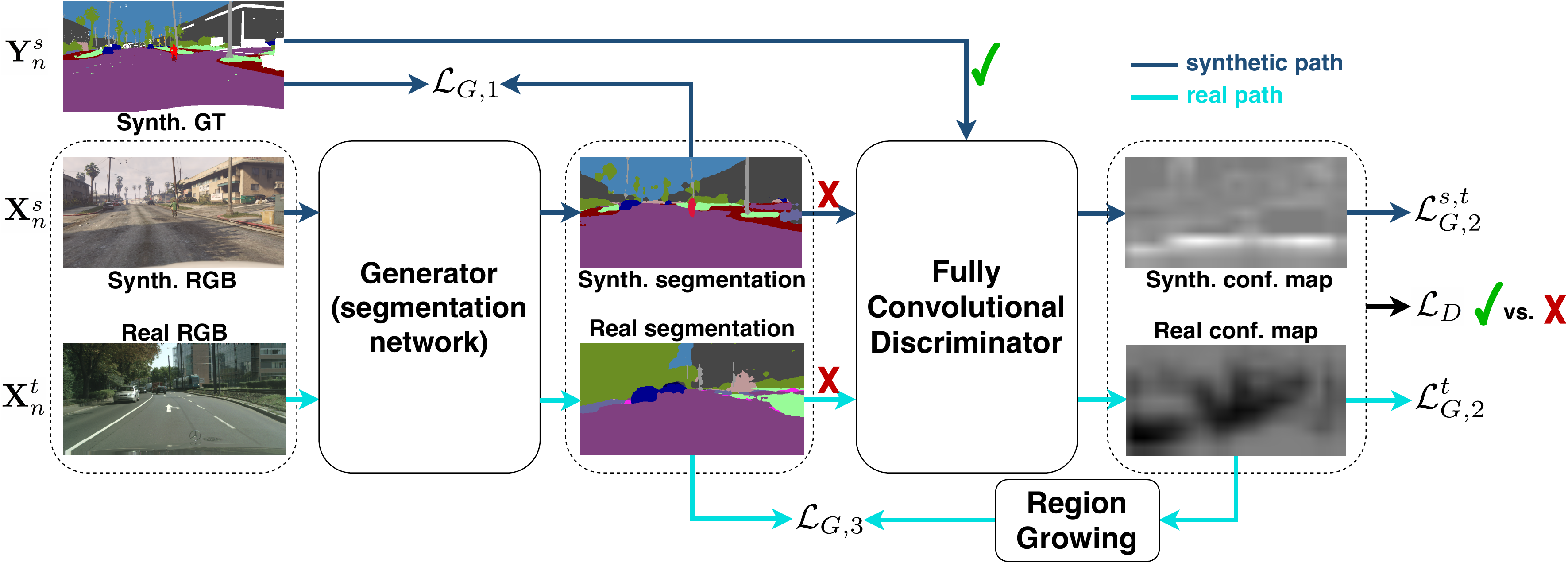}
\caption{\reviewres{Architecture of the proposed framework. The optimization is guided by a discriminator loss and  3 losses for the generator: a standard cross-entropy loss on synthetic data ($\mathcal{L}_{G,1}$), an adversarial loss ($\mathcal{L}^{s,t}_{G,2}$) and  a self-teaching loss for unlabeled real data ($\mathcal{L}_{G,3}$). \textit{Best viewed in colors. }}
}
\label{fig:scheme}
\end{figure*}

Our target is to train a semantic segmentation network in a supervised way on synthetic data and to adapt it in a unsupervised way to real data. \reviewtxt{In this paper, we name  this network $G$,  since it has the role of the \textit{generator} in the proposed adversarial training framework.} A supplementary discriminator network $D$ is used to evaluate the reliability of $G$'s output. This information can be employed to guide the adaptation of $G$ to unlabeled real data. 
In this section, we detail the CNN architectures and the training procedure implementing the unsupervised domain adaptation. Our approach is agnostic to the architecture of $G$ and in general any semantic segmentation network can be used\reviewtxt{. However,} in our experiments $G$ is a Deeplab v2 network \cite{chen2018deeplab}. This widely used model is based on the ResNet-101 backbone whose weights were pre-trained \cite{nekrasov}  on the MSCOCO dataset \cite{lin2014microsoft}.

Figure \ref{fig:scheme} shows the architecture of the proposed training framework. The optimization of the network is driven by the minimization of three loss functions. 
The first loss function ($\mathcal{L}_{G,1}$) is a standard multi-class cross-entropy. The segmentation network $G$ is trained to estimate for each input pixel the probability that it belongs to a class $c$ inside the set of possible classes $\mathcal{C}$. It is optimized only on labeled synthetic data since the ground truth is required.  
\reviewtxt{In the following, $G(\mathbf{X}^s_n)$ is used to represent the output of the segmentation network on the $n$-th input image, $\mathbf{X}^s_n$, from the source (synthetic) domain. $\mathbf{Y}^s_n$ is used to refer to the one-hot encoded ground truth segmentation related to input $\mathbf{X}^s_n$. In this scenario, the multi-class cross-entropy loss $\mathcal{L}_{G,1}$ is formulated as:}

\begin{equation}
\mathcal{L}_{G,1} = -  \sum_{p\in \mathbf{X}^s_n} \sum_{c \in \mathcal{C}} {\mathbf{Y}^s_n}^{(p)}[c] \cdot\log\big(G(\mathbf{X}^s_n)^{(p)}[c]\big)
\label{eq:L_G1}
\end{equation}

where $p$ is the index of a pixel in the considered image, $c$ is a specific class belonging to $\mathcal{C}$ and ${\mathbf{Y}^s_n}^{(p)}[c]$ and $G(\mathbf{X}^s_n)^{(p)}[c]$ are  the values relative to pixel $p$ and class $c$ respectively in the ground truth and in the generator ($G$) output.
As mentioned above, this loss can be computed only on the source (synthetic) domain where the semantic ground truth is available. 

The second and the third loss functions, minimized during the training of $G$, aim at adapting the semantic segmentation CNN $G$ to real data without using ground truth labels for real data.
These loss functions are implemented by means of the discriminator network $D$, that is trained to distinguish segmentation maps produced by the generator from the ground truth ones. The peculiarity of this discriminator network is that it produces a per-pixel estimation, differently from traditional adversarial frameworks where the discriminator outputs a single binary value for the whole input image.
The discriminator $D$ is made of a stack of $5$ convolutional layers each with $4\times 4$ kernels with a stride of $2$ and Leaky ReLU activation function. The number of filters (from the first layer to the last one) is $64$, $64$, $128$, $128$, $1$ and the cascade is followed by a bilinear upsampling to match the original input image resolution.
The discriminator is trained by minimizing the loss function $\mathcal{L}_{D}$, that  
is a standard cross-entropy loss, between $D$'s output and 
the one-hot encoding  indicating if the input is produced by $G$ (class $0$) or if it is the ground truth one-hot encoding semantic  segmentation (class $1$).  $\mathcal{L}_D$ can be formulated as:
\begin{equation}
\mathcal{L}_{D} = -\sum_{p\in \mathbf{X}^{s,t}_n}  \log(1-D(G(\mathbf{X}^{s,t}_n))^{( p )}) + \log(D(\mathbf{Y}^{s}_n)^{( p )})
\label{eq:L_D}
\end{equation}



Notice that the class $0$, associated to $G$'s output,  can be produced  both from an input $\mathbf{X}^s_n$ coming from the source domain and from a real world input $\mathbf{X}^t_n$. This means that $D$ can be trained on both synthetic and real data, trying to discriminate generated data from ground truth one.
The segmented source and target datasets share a similar statistic, since low level features of the color images are \reviewtxt{processed to leave place to the class statistic: for this} reason the training of $D$ on real and synthetic data is possible. 
Another possible issue in the training procedure could be related to the well distinguishable Dirac distributed segmentation ground truth data.  \reviewtxt{In principle, this could be easily distinguished from data produced by $G$. However, we have investigated this issue and in general $G$ produces segmentation maps very close to the Dirac distribution after a few training steps. This forces $D$  to capture also other statistical properties of the two different types of  input data.} Notice that this issue has been  investigated also in \cite{hung2018,biasetton2019} with similar conclusions.
The discriminator $D$ is used to implement the second loss function for the training of $G$, $\mathcal{L}_{G,2}^{s,t}$. This loss function is an adversarial loss since $G$, the generator in the traditional adversarial training scheme, is updated in order to create an output that has to look similar to ground truth data from the $D$ viewpoint. On a generic image $\mathbf{X}^{s,t}_n$ this loss function can be formulated as:

\begin{equation}
\mathcal{L}^{s,t}_{G,2} = - \sum_{p\in \mathbf{X}^{s,t}_n}   \log(D(G(\mathbf{X}^{s,t}_n))^{(p)})
\label{eq:L_G2}
\end{equation}

As for the training of $D$ (Eq.~\ref{eq:L_D}), $\mathcal{L}^{s,t}_{G,2}$ can be optimized both on the source and on the target data. In case the input is coming from the source dataset, \reviewtxt{we will refer to the loss function of Eq.~\ref{eq:L_G2} with $\mathcal{L}^{s}_{G,2}$, otherwise we will refer to it with $\mathcal{L}^{t}_{G,2}$ in case of target data as input.}
Notice that the generator is forced to adapt to the target real domain in an unsupervised way by minimizing $\mathcal{L}^{t}_{G,2}$. $G$ is forced to produce data similar to what $D$ considers ground truth also on real data. \reviewtxt{Remember that the ground truth is not used for this loss.}

The third loss function \reviewtxt{is inspired} from the work of Hung et al. \cite{hung2018}. The idea is to interpret the output of the discriminator $D$ as a measure of the reliability of the output of $G$ in case of synthetic and real data. This reliability  measure is used to realize a  self-training on real data.  The predictions of $G$, assumed to be reliable by $D$, are converted to the one-hot encoding and are used as a  self-taught ground truth to train $G$ on unlabeled target real data. This loss can be formulated as


\begin{equation}
\label{eq:L_G3}
\mathcal{L}_{G,3} \!= \!-\! \!\! \! \sum_{p\in \mathbf{X}_n^t} \!
\sum_{c \in \mathcal{C}}
\! D_{\!R}(\mathbf{X}_n^t)^{(p)} \! \cdot
W_c^s \cdot \, 
\!\mathbf{\hat{Y}}_n^{(p)}[c] \cdot
\log\!\big(G(\mathbf{X}_n^t)^{(p)}[c]\big)
\end{equation}
where $\mathbf{\hat{Y}}_n$ is the one-hot encoded ground truth derived from the per-class argmax of the generated probability map $G(\mathbf{X}_n)$.
Each contribution to the loss is weighted by two terms. The first ($D_R$) is a weighting term dependent on the  output of the discriminator refined by a region growing procedure that exploits pixel aggregation to improve the confidence estimation.
 The second ($W_c^s$) is a weighting function proportional to the class \reviewtxt{frequencies} on the source domain.


More in detail, the first term \reviewtxt{finds} the reliable locations \reviewtxt{in the segmented map and assigns to them a weight interpreted as a confidence measure. The module computing this weighting mask is named $D_R(\cdot)$ and it takes as input a real image $\mathbf{X}_n^t$. In the first step, a mask $m_{T_u}$ is computed selecting confident points by applying a threshold $T_u$ to the output of the discriminator with input $G(\mathbf{X}_n^t)$. The discriminator output is interpreted as a confidence map related to the segmentation map estimated on $\mathbf{X}_n^t$ in this phase. Formally, at each pixel location $p$ we have:}

\begin{equation}
m_{T_u}^{(p)}= 
\begin{cases} 
1 & \mathrm{if} \: D(G(\mathbf{X}_n^t))^{(p)}>T_u \\
0 & \mathrm{otherwise} \\
\end{cases}
\label{eq:mask}
\end{equation}

In the second step, for a generic confident pixel $p$ in  $m_{T_u}$, \reviewtxt{assigned by G to class $c^*$,} the algorithm expands the confident region to a generic adjacent pixel $p'\in\mathbf{X}_n^t$ if the  output of the segmentation network  for the  class $c^*$ \reviewtxt{(i.e., the one selected for point $p$)}  is greater than a threshold $T_{\mathcal{R}}$ at location $p'$. \reviewtxt{More formally, $p'$ is added to the mask if} $G(\mathbf{X}_n^t)^{(p')}[c^*] > T_{\mathcal{R}}$. We will denote with $m^{R}_{T_u}$ the mask obtained by applying this region growing process to the original mask $m_{T_u}$.
Finally, for each location $p^R$ selected by the updated mask $m^{R}_{T_u}$ the weight  is given by the corresponding output of the discriminator $D(G(\mathbf{X}_n^t))^{(p^R)}$. Thus, the resulting weights $D_R(\mathbf{X}_n^t)$ are: 
\begin{equation}
D_R(\mathbf{X}_n^t)=m^{R}_{T_u}\cdot D(G(\mathbf{X}_n^t))
\end{equation}
i.e., the weight is equal to the discriminator output for points selected by $m^{R}_{T_u}$ and to $0$ for points not selected by the mask. Empirically we set $T_u=0.2$  and $T_{\mathcal{R}}=1-10^{-5}$ thus achieving high reliability when expanding the confidence map. 

The second weighting function is related to the class frequency on the source domain ($W_c^s$). \reviewtxt{ It is defined as:}
\begin{equation}
\label{eq:weight}
W^{t}_c =
\displaystyle
1-\frac{\sum_{n} \lvert {p \in \mathbf{X}^{s}_n} \land p \in c \rvert}
{\sum_{n} \lvert {p \in \mathbf{X}^{s}_n} \rvert } ,
\end{equation}
where $\lvert\ \cdot\ \rvert$ represents the cardinality of the considered set.

This weighting function balances the overall loss when unlabeled data of the target set are used, avoiding that rare and tiny objects (e.g., \textit{traffic lights} or \textit{pole}) are forgotten and replaced by more frequent and large ones (such as \textit{road}, \textit{building}). Notice that $W_c^s$ is estimated on source data since the ground truth of the target data is assumed to be unknown during the training phase. Furthermore, $W_c^s$ does not change during the training process and so it is computed only once.

Finally, the overall loss function for the training of $G$ is a weighted average of the three losses, i.e.:
\begin{equation}
\label{eq:L_full}
\mathcal{L}_{full} = \mathcal{L}_{G,1} + w^{s,t}\mathcal{L}_{G,2}^{s,t} + w' \mathcal{L}_{G,3}
\end{equation}
\reviewres{We empirically set the weighting parameters as specified in Section~\ref{sec:results}.}
The discriminator is trained minimizing $\mathcal{L}_{D}$ (Eq.~\ref{eq:L_D}) on ground truth labels and on the generator output computed on a mixed batch composed by both source and target data. During the first $5000$ steps, the loss $\mathcal{L}_{G,3}$ is disabled, setting $w'=0$, allowing the discriminator to learn how to produce higher quality confidence maps before using them. After this initial phase, all the three components of the loss are enabled and the training ends after $20000$ steps.
\section{Datasets}
\label{sec:datasets}

The proposed unsupervised domain adaptation framework has been trained and evaluated using 4 different datasets. 
Recall that the  target is to  train the semantic segmentation network using labeled synthetic road scenes while no labels are available for real world data.
The supervised synthetic training exploits two publicly available datasets, i.e., GTA5 \cite{Richter2016} and SYNTHIA \cite{ros2016}. 
The real world datasets used for the unsupervised adaptation and for the result evaluation are instead Cityscapes \cite{Cordts2016} and Mapillary \cite{neuhold2017mapillary}. 
Notice that the evaluation scenario is the same of recent competing approaches  as \cite{hoffman2016, sankaranarayanan2018, zhang2017} in order to allow for a fair comparison. During the training stage all the images have been resized and cropped to $750\times 375$ $\mathrm{px}$ for memory constraints. The testing on the real datasets, instead, has been carried out at their original resolution.

The \textbf{GTA5} dataset \cite{Richter2016} contains $24966$  synthetic images with pixel level semantic annotation. The images have been rendered using the open-world video game \textit{Grand Theft Auto 5} and are all from the car perspective in the streets of American-style virtual cities. 
 The images  have an impressive visual quality and are very realistic since they come  from a high budget commercial production. We used $23966$ images for the supervised \reviewtxt{training, while} $1000$ have been taken out for validation purposes. 
 There are 19 semantic classes which are compatible with the ones of the exploited real world datasets. 

The  \textit{SYNTHIA-RAND-CITYSCAPES} subset of the \textbf{SYNTHIA} dataset \cite{ros2016} contains  $9400$ synthetic $1280 \times 760$ $\mathrm{px}$ images with pixel level semantic annotation.  
The images have been rendered with  an ad-hoc engine, allowing to obtain a large variability of street scenes. \reviewtxt{In this case they come from virtual European-style towns in different environments under various light and weather conditions.}
On the other hand,  the visual quality is not the same of the commercial video game GTA5. 
The semantic labels are compatible with $16$ of the $19$ classes of Cityscapes (for the evaluation  on the Cityscapes dataset, only the $16$ classes contained in both  datasets are taken into consideration). 
We used $9300$ images for the supervised training while  $100$ have been taken out for validation purposes. 

The \textbf{Cityscapes} dataset \cite{Cordts2016} contains $2975$ color images \reviewtxt{with pixel level semantic annotation. The images have a resolution of $2048 \times 1024$ $\mathrm{px}$ and have been captured on the streets of $50$ European cities.  
We used the labels only for experimental evaluation, since the domain adaptation procedure is unsupervised.}
The original training set (without the labels) has been used for unsupervised adaptation, while the $500$ images in the original validation set have been used as a test set (as done  by  competing approaches since the test set labels are not available).

The \textbf{Mapillary} dataset \cite{neuhold2017mapillary} contains  $20000$ high resolution images  
taken from different cameras in many different locations. 
The variability in classes, appearance, acquisition settings and geo-localization makes the dataset the most complete and of highest quality in the field.
As for Cityscapes we used this  dataset for unsupervised domain adaptation and testing. 
The semantic annotations have been re-conducted  to the labels of the Cityscapes dataset following the mapping in \cite{kim2018attribute}. 
We exploited the $18000$  training images (without the labels) for unsupervised training and the $2000$ images in the original validation set as test set (as done by competing approaches). 

\begin{table*}[htbp]
\setlength{\tabcolsep}{1.6pt}
\centering
\begin{tabular}{|c|c|c|c|c|c|}
\hline
  & \textbf{Hoffman et al. \cite{hoffman2016}} &  \textbf{Hung et al. \cite{hung2018}} & \textbf{Zhang et al. \cite{zhang2017}}  & \textbf{Biasetton et al. \cite{biasetton2019}} & \textbf{Ours} \\\hline
  
  Domain adaptation strategy & \makecell{Adversarial feature alignment,\\ label distributions} &  \makecell{Adversarial learning\\and self-teaching} & \makecell{Labels distribution\\(global and on superpixels)} & \makecell{Adversarial learning\\and self-teaching} & \makecell{Adversarial learning\\and self-teaching} \\\hline
  
  \textit{D} input & \makecell{Source features vs.\\target features} & \makecell{Ground truth vs.\\predicted maps} & - & \makecell{Ground truth vs.\\predicted maps} & \makecell{Ground truth vs.\\predicted maps} \\\hline
  
  \textit{D} output & Binary & Confidence map & - & Confidence map& Confidence map \\\hline
  
  \textit{G} backbone & FCN-8s & Deeplab v2 & FCN-8s & Deeplab v2 & Deeplab v2 \\\hline
  
  Loss & CE, ADV-Feat., LD & CE, ADV, ST & CE, superpixel, LD & CE, ADV, ST & CE, ADV, ST \\\hline
  
  Self-teaching & No & Yes & No & Yes & \makecell{Yes, with soft selection\\and region growing} \\\hline
  
  Class-weighting & Yes & No & No & Yes & Yes \\\hline
  
  Pre/post processing & Label distribution and superpixel & - & Superpixel segmentation  & - & - \\\hline
\end{tabular}
\vspace{0.1cm}
\caption{\reviewres{Summary of compared methodologies. Loss components are expressed as  CE: cross entropy,  ADV: adversarial loss,  LD: labels distribution,  ST: self-teaching.}}
\label{tab:comparison}
\end{table*}

\begin{table*}[htbp]
\setlength{\tabcolsep}{1.6pt}
\centering
\begin{tabular}{|c|ccccccccccccccccccc|c|}
\hline
\textbf{GTA5} $\rightarrow$ \textbf{Cityscapes}   & \rotatebox{90}{road} &  \rotatebox{90}{sidewalk} &  \rotatebox{90}{building} &  \rotatebox{90}{wall} &\rotatebox{90}{fence} & \rotatebox{90}{pole} & \rotatebox{90}{t light} 
  &\rotatebox{90}{t sign} & \rotatebox{90}{veg} & \rotatebox{90}{terrain} & \rotatebox{90}{sky} & \rotatebox{90}{person}& \rotatebox{90}{rider} & \rotatebox{90}{car} 
  & \rotatebox{90}{truck} & \rotatebox{90}{bus} & \rotatebox{90}{train} &  \rotatebox{90}{mbike} & \rotatebox{90}{bike} & \rotatebox{90}{mean} \\
 \hline
Supervised ($\mathcal{L}_{G,1}$ only) & 45.3 & 20.6 & 50.1 & 9.3 & 12.7 & 19.5 & 4.3 & 0.7 & 81.9 & 21.1 & 63.3 & 52.0 & 1.7 & 77.9 & 26.0 & 39.8 & 0.1 & 4.7 & 0.0 & 27.9\\
Ours (full) & 81.0 & 19.6 & 65.8 & \textbf{20.7} & \textbf{12.9} & \textbf{20.9} & 6.6 & 0.2 & 82.4 & \textbf{33.0} & \textbf{68.2} & \textbf{54.9} & 6.2 & 80.3 & 28.1 & 41.6 & \textbf{2.4} & 8.5 & 0.0 & \textbf{33.3} \\\hline
Hoffman et al. \cite{hoffman2016} & 70.4 &  \textbf{32.4} & 62.1 & 14.9 & 5.4 & 10.9 & 14.2 &  2.7 &   79.2 &  21.3 & 64.6 & 44.1 &  4.2&  70.4 & 8.0 & 7.3 & 0.0 & 3.5 & 0.0 & 27.1\\
Hung et al. \cite{hung2018} & \textbf{81.7} & 0.3 & 68.4 & 4.5 & 2.7 & 8.5 & 0.6 & 0.0 & \textbf{82.7} & 21.5 & 67.9 & 40.0 & 3.3 & \textbf{80.7} & \textbf{34.2} & \textbf{45.9 }& 0.2 & 8.7 & 0.0 & 29.0\\
Zhang et al. \cite{zhang2017} & 74.9 & 22.0 & \textbf{71.7} & 6.0 & 11.9 & 8.4 & \textbf{16.3} & \textbf{11.1} & 75.7 & 13.3 & 66.5 & 38.0 & \textbf{9.3} & 55.2 & 18.8 & 18.9 & 0.0 & \textbf{16.8} & \textbf{14.6} & 28.9 \\
Biasetton et al. \cite{biasetton2019} & 54.9 & 23.8 & 50.9 & 16.2 & 11.2 & 20.0 & 3.2 & 0.0 & 79.7 & 31.6 & 64.9 & 52.5 & 7.9 & 79.5 & 27.2 & 41.8 & 0.5 & 10.7 & 1.3 & 30.4\\
\hline
\end{tabular}
\vspace{0.1cm}
\caption{mIoU on the different classes of the Cityscapes validation set. The approaches have been trained in a supervised way on the GTA5 dataset and the unsupervised domain adaptation has been performed using the Cityscapes training set. \reviewres{The highest value in each column is highlighted in bold.}}
\label{tab:GTA}

\vspace{0.5cm}

\setlength{\tabcolsep}{3.6pt}
\centering
\begin{tabular}{|c|cccccccccccccccc|c|}
\hline
\textbf{SYNTHIA} $\rightarrow$ \textbf{Cityscapes}  & \rotatebox{90}{road} &  \rotatebox{90}{sidewalk} &  \rotatebox{90}{building} &  \rotatebox{90}{wall} &\rotatebox{90}{fence} & \rotatebox{90}{pole} & \rotatebox{90}{t light} 
  &\rotatebox{90}{t sign} & \rotatebox{90}{veg} & \rotatebox{90}{sky} & \rotatebox{90}{person}& \rotatebox{90}{rider} & \rotatebox{90}{car} 
 & \rotatebox{90}{bus} &  \rotatebox{90}{mbike} & \rotatebox{90}{bike} & \rotatebox{90}{mean} \\
 \hline
Supervised ($\mathcal{L}_{G,1}$ only) & 10.3 & 20.5 & 35.5 & 1.5 & 0.0 & \textbf{28.9} & 0.0 & 1.2 & 83.1 & 74.8 & \textbf{53.5} & 7.5 & 65.8 & 18.1 & \textbf{4.7} & 1.0 & 25.4  \\
Ours (full) & \textbf{80.7} & 0.3 & \textbf{75.0} & 0.0 & 0.0 & 19.5 & 0.0 & 0.4 & 84.0 & \textbf{79.4} & 46.6 & 0.8 & \textbf{80.8} & \textbf{32.8} & 0.5 & 0.5 & \textbf{31.3}  \\\hline
Hoffman et al. \cite{hoffman2016} & 11.5 & 19.6 & 30.8 & \textbf{4.4} & 0.0 & 20.3 & 0.1 & \textbf{11.7} & 42.3 & 68.7 & 51.2 & 3.8 & 54.0 &  3.2 & 0.2 & 0.6 & 20.1  \\
Hung et al. \cite{hung2018} & 72.5 & 0.0 & 63.8 & 0.0 & 0.0 & 16.3 & 0.0 & 0.5 & \textbf{84.7} & 76.9 & 45.3 & 1.5 & 77.6 & 31.3 & 0.0 & 0.1 & 29.4 \\
Zhang et al. \cite{zhang2017} & 65.2 & \textbf{26.1} & 74.9 & 0.1 & \textbf{0.5} & 10.7 & \textbf{3.7} & 3.0 & 76.1 & 70.6 & 47.1 & \textbf{8.2} & 43.2 & 20.7 & 0.7 & \textbf{13.1} & 29.0 \\
Biasetton et al. \cite{biasetton2019} & 78.4 & 0.1 & 73.2 & 0.0 & 0.0 & 16.9 & 0.0 & 0.2 & 84.3 & 78.8 & 46.0 & 0.3 & 74.9 & 30.8 & 0.0 & 0.1 & 30.2  \\
\hline
\end{tabular}
\vspace{0.1cm}
\caption{mIoU on the different classes of the Cityscapes validation set. The approaches have been trained in a supervised way on the SYNTHIA dataset and the unsupervised domain adaptation has been performed using the Cityscapes training set. \reviewres{The highest value in each column is highlighted in bold.}}
\label{tab:SYNTHIA}
\end{table*}

\section{Experimental Results}
\label{sec:results}

The target of the proposed approach is to adapt a deep network trained on synthetic data to real world scenes. To evaluate the performance on this task we used the 4 different datasets introduced in Section \ref{sec:datasets}.
We started by evaluating the performance on the validation set of Cityscapes. 
In the first experiment, we trained the network using the scenes from the GTA5  dataset to compute the supervised loss $\mathcal{L}_{G,1}$ and the adversarial loss $\mathcal{L}_{G,2}^s$ while the training scenes of the Cityscapes dataset have been used for the unsupervised domain adaptation, i.e., to compute the losses $\mathcal{L}_{G,2}^t$ and $\mathcal{L}_{G,3}$. Notice that no labels from the Cityscapes training set have been used.  
In the second experiment, we performed the same procedure but we replaced the GTA5 dataset with the SYNTHIA one.

Then\reviewtxt{,} we switched to the Mapillary dataset and we repeated the two experiments using this dataset: we performed the supervised training with GTA\reviewtxt{5} or SYNTHIA and we used the training set of Mapillary, \reviewtxt{without any label,} for the unsupervised domain adaptation. \reviewtxt{Similarly to the previous scenario,} we evaluated the results on the validation split of Mapillary.

The proposed deep learning scheme has been implemented using the TensorFlow framework. 
The  generator network $G$ \reviewtxt{(we used a Deeplab v2 model)} has been trained as proposed in \cite{chen2018deeplab} using the Stochastic Gradient Descent (SGD)  optimizer with momentum set to $0.9$ and weight decay to $10^{-4}$. 
The discriminator $D$ has been trained using the Adam optimizer. The learning rate employed for both $G$ and $D$ started from $10^{-4}$ and was decreased up to $10^{-6}$ by means of a polynomial decay with power $0.9$. 
We trained the two networks for $20000$ iterations 
on a NVIDIA GTX 1080 Ti GPU. The longest training inside this work, i.e., the one with all the loss components enabled, takes about 20 hours to complete. Further resources and the code of our approach are available at: \url{https://lttm.dei.unipd.it/paper_data/semanticDA}.

\reviewres{We assess the quality of our approach computing the mean Intersection over Union (mIoU), as done by all competing approaches. Moreover, we compared our results with some recent frameworks \cite{hoffman2016, zhang2017, hung2018, biasetton2019}, whose main design choices are briefly summarized in Table~\ref{tab:comparison}.}

\subsection{Evaluation on the Cityscapes Dataset}

We started the experimental evaluation from the Cityscapes dataset. The \reviewtxt{performance} have been computed by comparing the predictions on the Cityscapes validation set with the ground truth.  
%
Table \ref{tab:GTA} refers to the first experiment (i.e., using GTA5 for the supervised training). It shows the accuracy obtained with standard supervised training, with the proposed approach 
and with some state-of-the-art approaches.
By simply training the network in a supervised way on the GTA5 dataset and then performing inference on real world data from the Cityscapes dataset  a mIoU of $27.9\%$ can be obtained.
The proposed unsupervised domain adaptation strategy allows to enhance the accuracy to $33.3\%$ with an improvement of $5.4\%$.
By looking more in detail to the various class accuracies, it is possible to see that the accuracy has increased on almost all the classes (only on two of them the accuracy has slightly decreased)\reviewtxt{. In particular,} there is a large improvement on the most common classes corresponding to large structures, since the domain adaptation strategy allows to better learn their statistics in the new domain. At the same time the performance improves also on less frequent classes corresponding to small objects \reviewtxt{ due to the usage of the class weights $W_c^t$ in the self-teaching loss component.} 

\reviewtxt{The method of Hung et al. \cite{hung2018}, based on a similar framework, achieves a lower accuracy of $29\%$ mostly because it struggles with small structures and uncommon classes.
The methods in \cite{zhang2017, hoffman2016} also have lower performance; however, they are also based on a different generator network.}
The older version of our method, introduced in \cite{biasetton2019}, achieves an accuracy of $30.4\%$, with a gap of almost $3\%$ w.r.t. the proposed approach, proving that the newly introduced elements (i.e., the weighting in the self-teaching and the region growing strategy) have a relevant impact on the performance.

 Figure \ref{fig:qual_res}a shows the output of the supervised training, of the methods of \cite{hung2018} and \cite{biasetton2019} and of our approach on some sample scenes, using the GTA5 dataset as source dataset and the Cityscapes as target one. The supervised training leads to reasonable results, but some small objects get lost or the object contours are badly captured (e.g., the rider in row 1 or the poles in row 3). Furthermore, some regions of the street  are corrupted by noise (e.g., see rows 1 and 2). 
The approach of \cite{hung2018} seems to lose some structures (e.g., the terrain in the third row) and presents issues with small objects (the poles in row 3 get completely lost) as pointed out before. 
The old version of the approach \cite{biasetton2019} has better performance, \reviewtxt{ for example 
 the people are better preserved and the structures have better defined edges but there are still artifacts, e.g., the road surface in row 2 and 3.}
Finally\reviewtxt{,} the proposed method has the best performance showing a good capability of detecting small objects and structures and at the same time a reliable recognition of the road and of the larger elements in the scene: \reviewtxt{ in all the selected images it obtains a cleaner representation of the road  removing the \textit{sidewalk} class where is not present but at the same time correctly localizes it in the second row differently from the other methods. Similar discussion holds  for the \textit{terrain} class in row 3 and for the \textit{pole} class whose detection has been highly improved w.r.t. \cite{hung2018}.}

\newcommand{\imgsize}{27.3mm}
\begin{figure*}[htbp]
\centering
\begin{subfigure}[htbp]{\textwidth}
\hspace{0.22cm}
\resizebox{0.965\textwidth}{!}{%
\begin{tabular}{cccccccccc}
\cellcolor[HTML]{804080}{\color[HTML]{FFFFFF} \textbf{road}} & \cellcolor[HTML]{F423E8}\textbf{sidewalk} & \cellcolor[HTML]{464646}{\color[HTML]{FFFFFF} \textbf{building}} & \cellcolor[HTML]{66669C}{\color[HTML]{FFFFFF} \textbf{wall}} & \cellcolor[HTML]{BE9999}\textbf{fence} & \cellcolor[HTML]{999999}\textbf{pole} & \cellcolor[HTML]{FAAA1E}\textbf{traffic light} & \cellcolor[HTML]{DCDC00}\textbf{traffic sign} &\cellcolor[HTML]{6B8E23} \textbf{vegetation} & \cellcolor[HTML]{98FB98}\textbf{terrain} \\ \cline{10-10} 
\cellcolor[HTML]{4682B4}\textbf{sky} & \cellcolor[HTML]{DC143C}{\color[HTML]{FFFFFF} \textbf{person}} & \cellcolor[HTML]{FF0000}{\color[HTML]{FFFFFF} \textbf{rider}} & \cellcolor[HTML]{00008E}{\color[HTML]{FFFFFF} \textbf{car}} & \cellcolor[HTML]{000046}{\color[HTML]{FFFFFF} \textbf{truck}} & \cellcolor[HTML]{003C64}{\color[HTML]{FFFFFF} \textbf{bus}} & \cellcolor[HTML]{005064}{\color[HTML]{FFFFFF} \textbf{train}} & \cellcolor[HTML]{0000E6}{\color[HTML]{FFFFFF} \textbf{motorcycle}} & \multicolumn{1}{c|}{\cellcolor[HTML]{770B20}{\color[HTML]{FFFFFF} \textbf{bicycle}}} & \multicolumn{1}{c|}{\textbf{unlabeled}} \\ \cline{10-10} 
\end{tabular}%
}
\vspace{0.1cm}
\end{subfigure}
\setlength{\tabcolsep}{1pt} 
\centering
\begin{subfigure}[htbp]{2\textwidth}
\begin{tabular}{c|c|c|cccccc}
\cline{2-3}
  
  \multirow{6}{*}{\vspace{-27ex}a)} & \multirow{6}{*}{\rotatebox{90}{\hspace{-20ex}To Cityscapes}} & \multirow{3}{*}{\rotatebox{90}{\hspace{-7ex}From GTA5}} &
  \includegraphics[width=\imgsize]{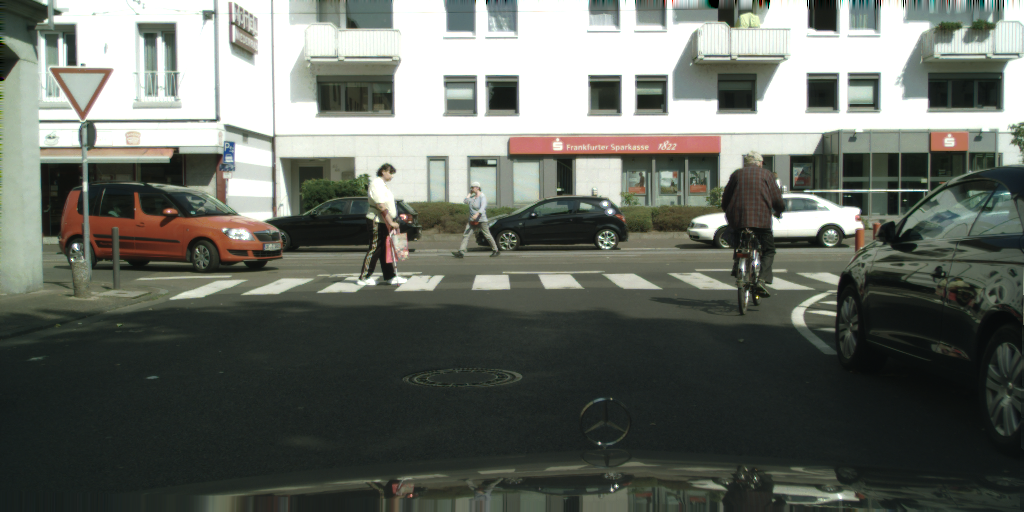} &
   \includegraphics[width=\imgsize]{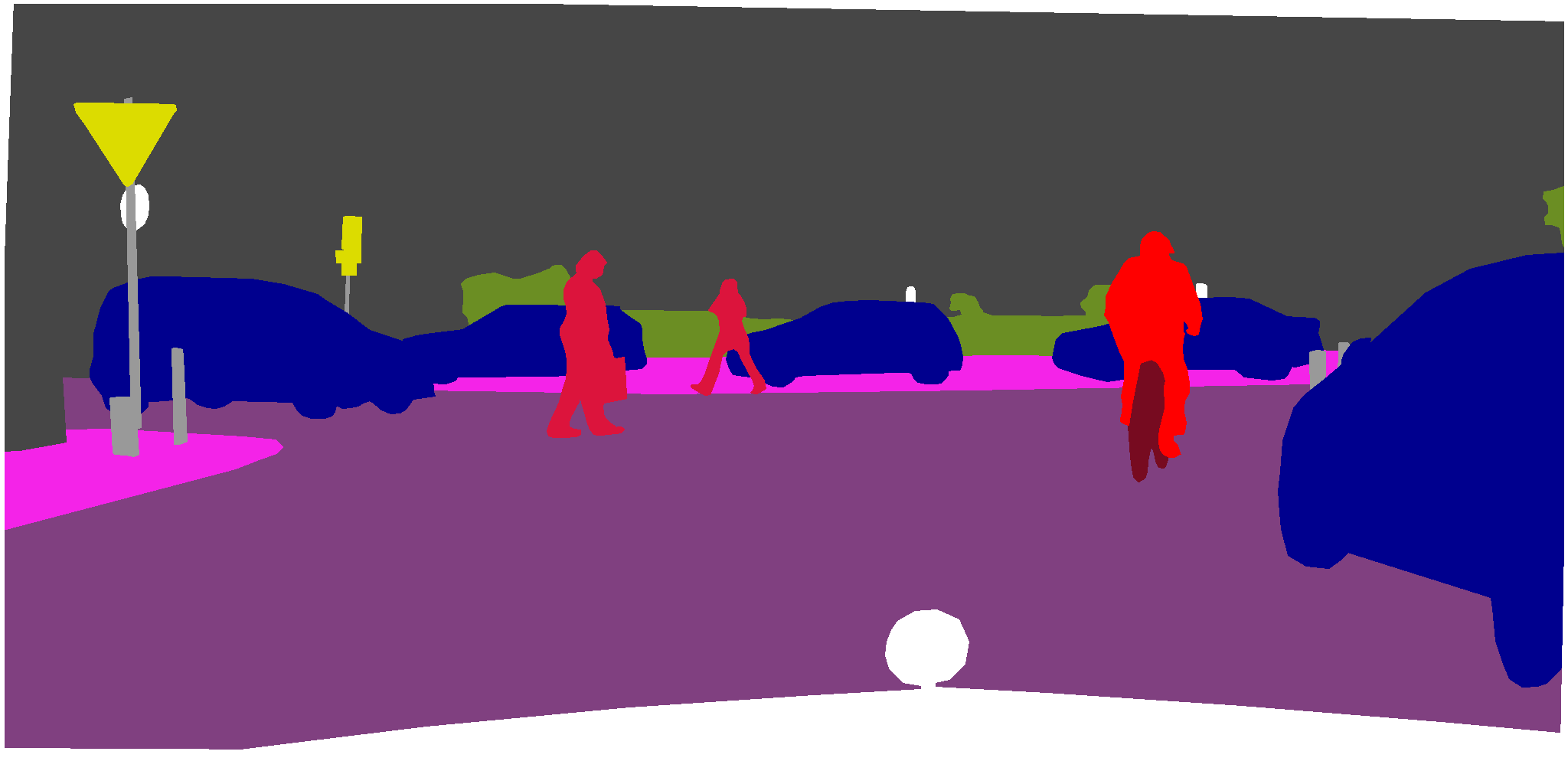} & 
   \includegraphics[width=\imgsize]{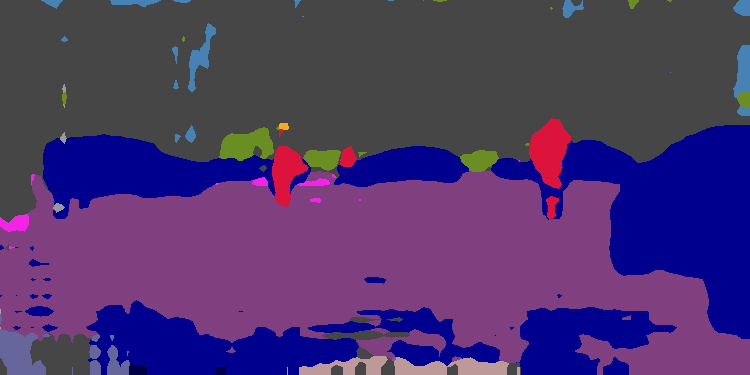} & 
   \includegraphics[width=\imgsize]{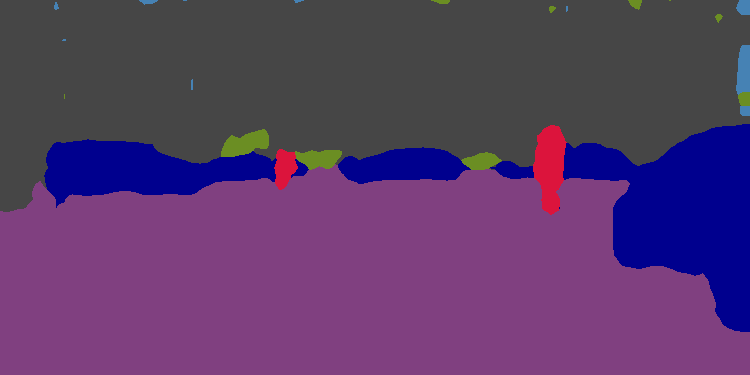} & 
   \includegraphics[width=\imgsize]{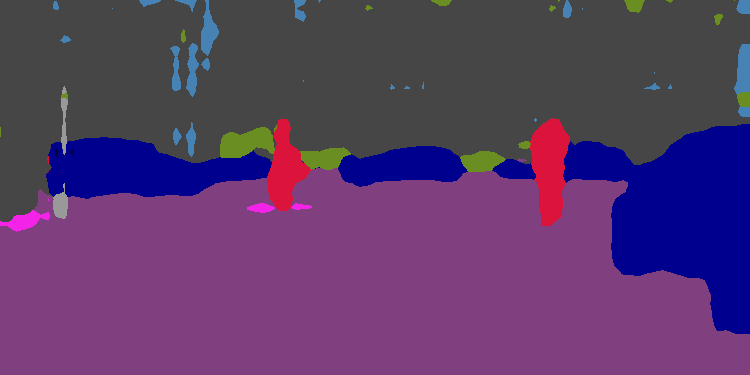} & \includegraphics[width=\imgsize]{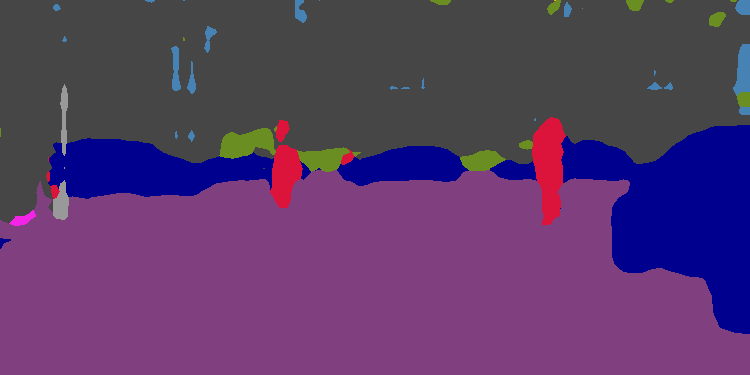} \\
  
&& & \includegraphics[width=\imgsize]{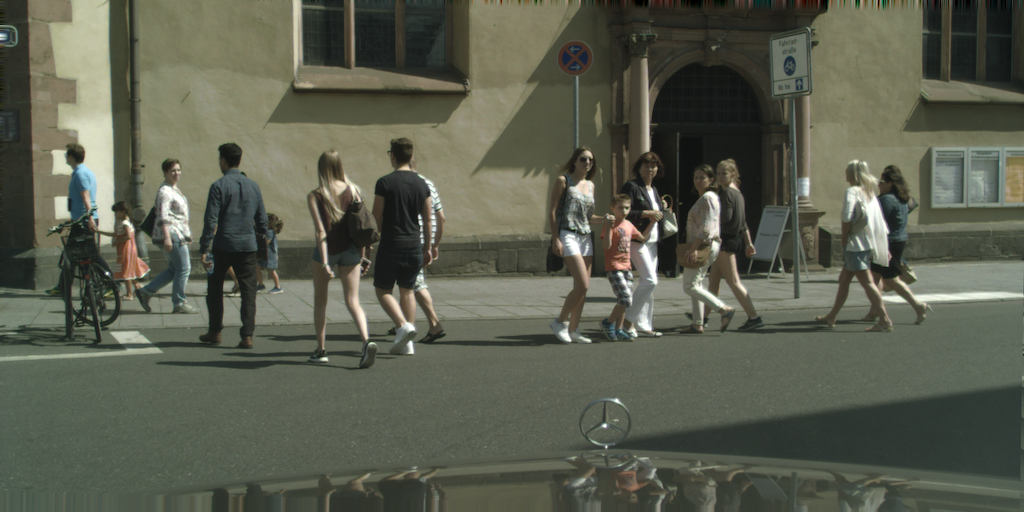} &
   \includegraphics[width=\imgsize]{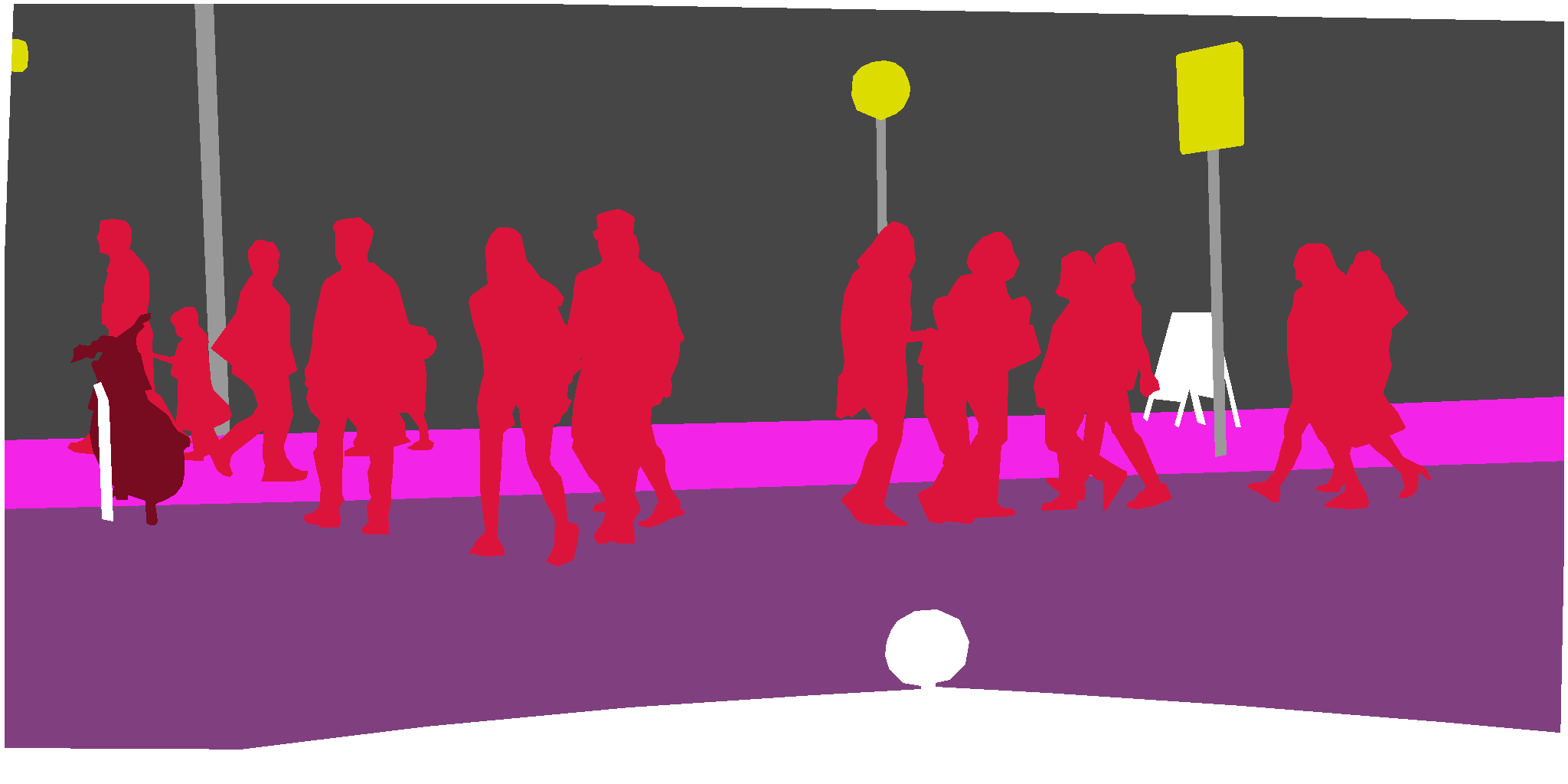} & 
   \includegraphics[width=\imgsize]{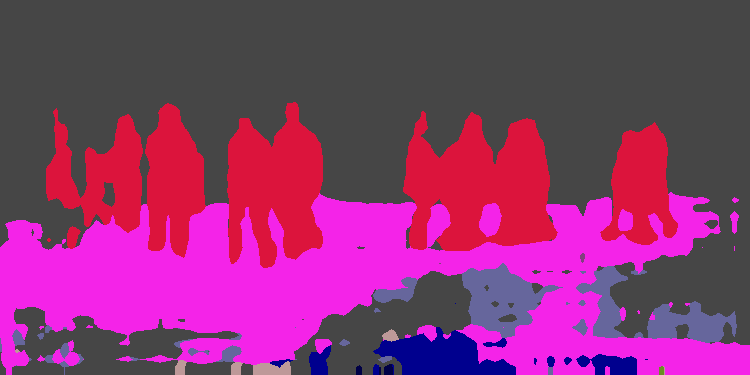} & 
   \includegraphics[width=\imgsize]{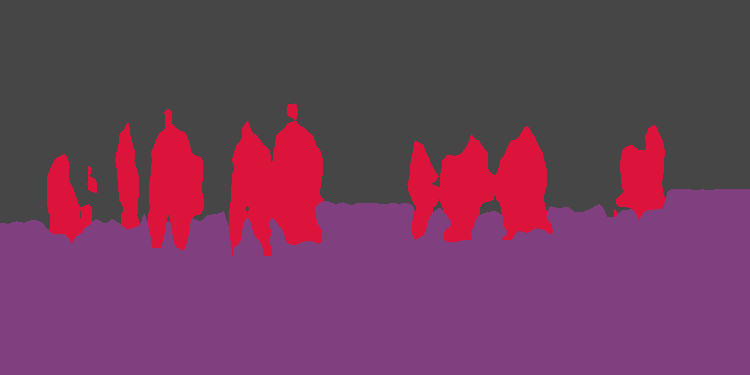} & 
   \includegraphics[width=\imgsize]{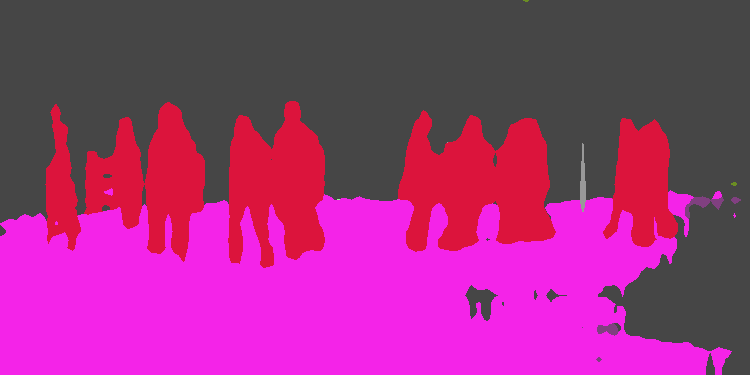} & 
   \includegraphics[width=\imgsize]{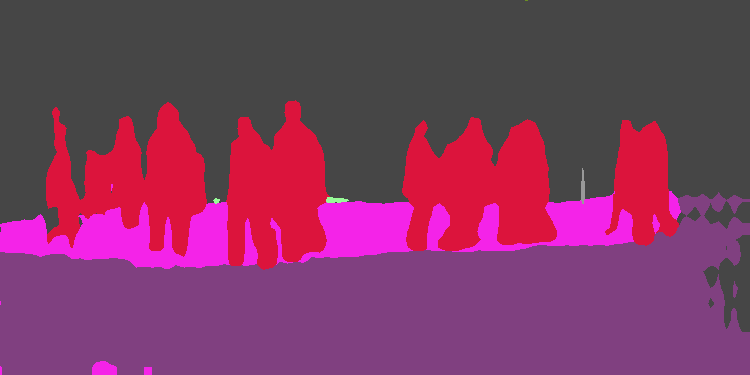} \\  
  
     && & \includegraphics[width=\imgsize]{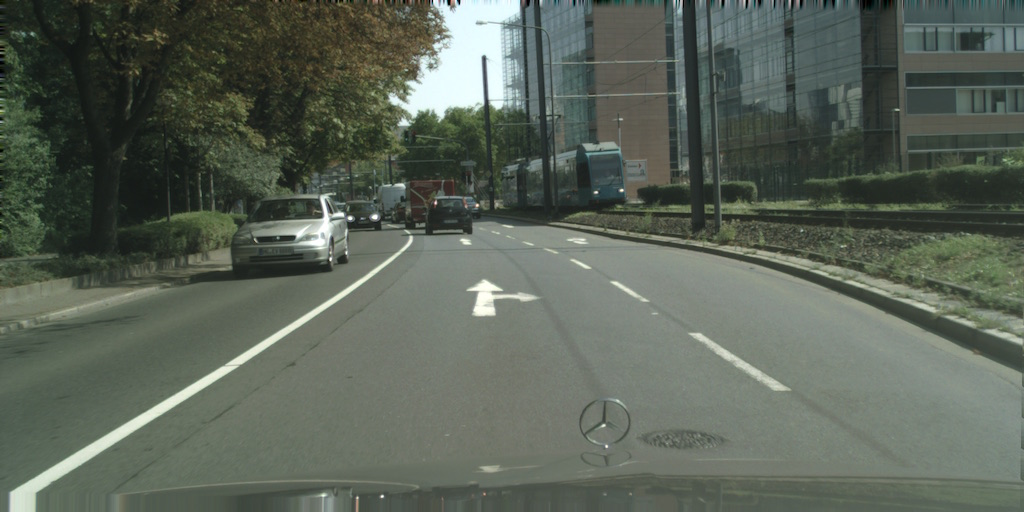} &
   \includegraphics[width=\imgsize]{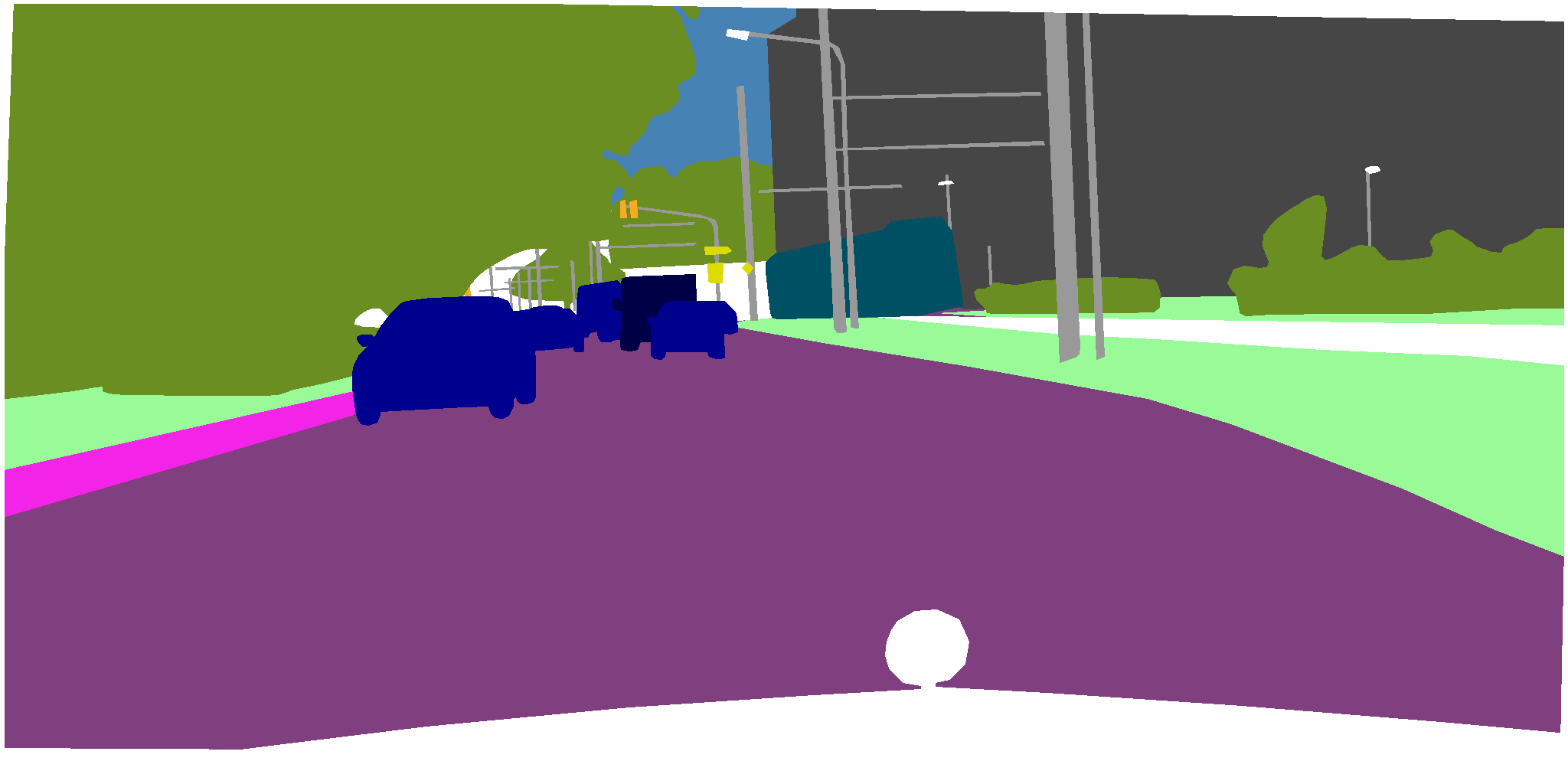} & 
   \includegraphics[width=\imgsize]{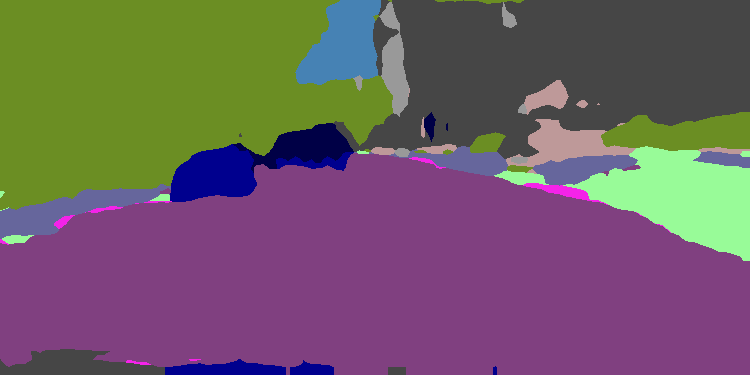} & 
   \includegraphics[width=\imgsize]{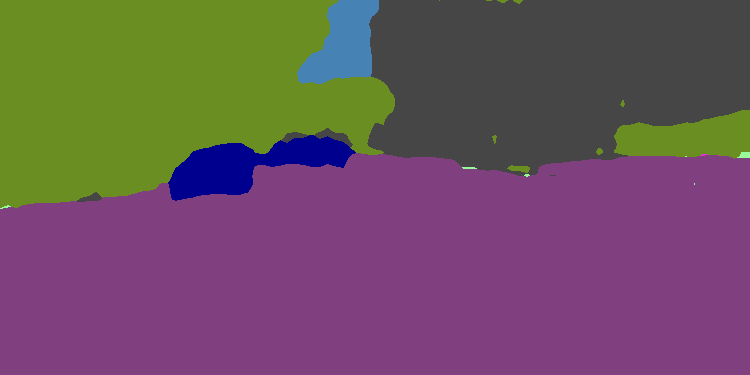} & 
   \includegraphics[width=\imgsize]{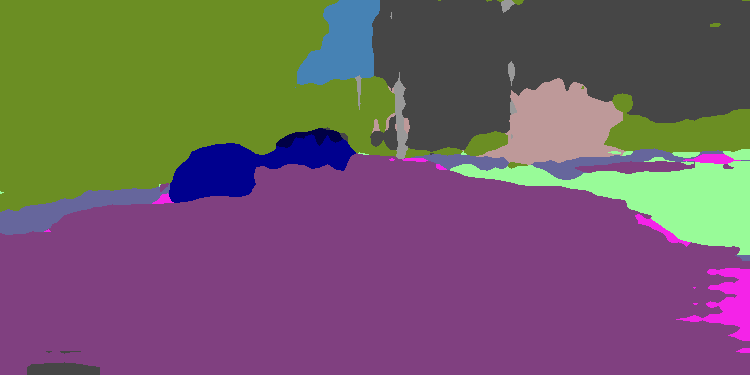} & 
   \includegraphics[width=\imgsize]{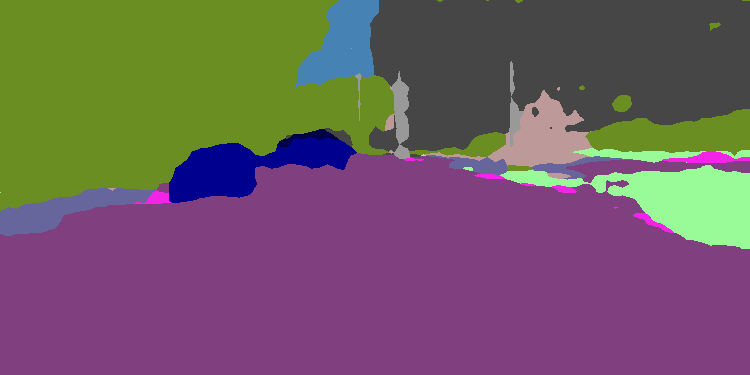} \\\cline{3-3}
  
   && \multirow{3}{*}{\rotatebox{90}{\hspace{+3ex}From SYNTHIA}} & \includegraphics[width=\imgsize]{images/DA_GTA-Cityscapes/rgb/frankfurt_000000_001016_leftImg8bit.png} &
  \includegraphics[width=\imgsize]{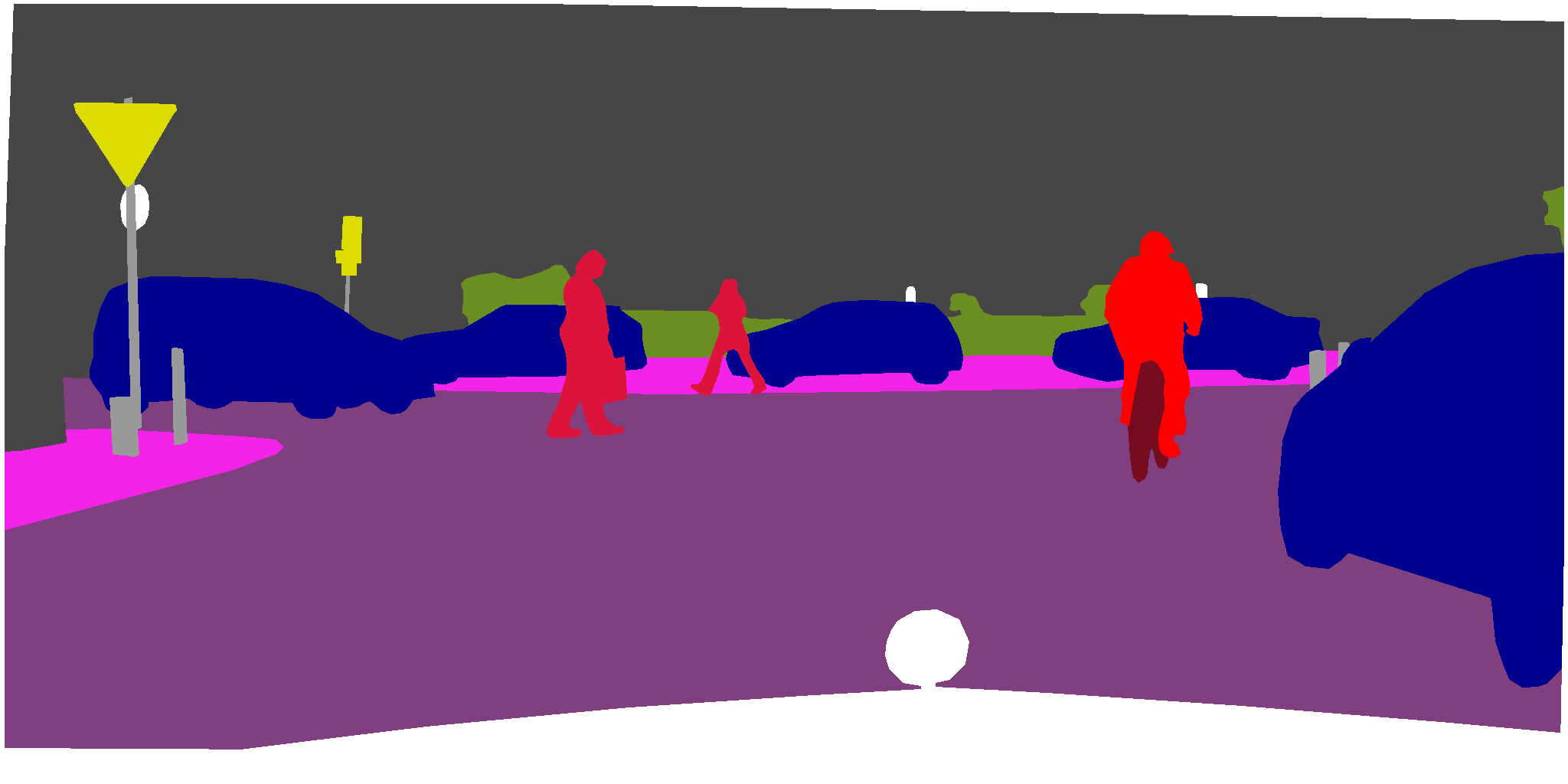} & 
  \includegraphics[width=\imgsize]{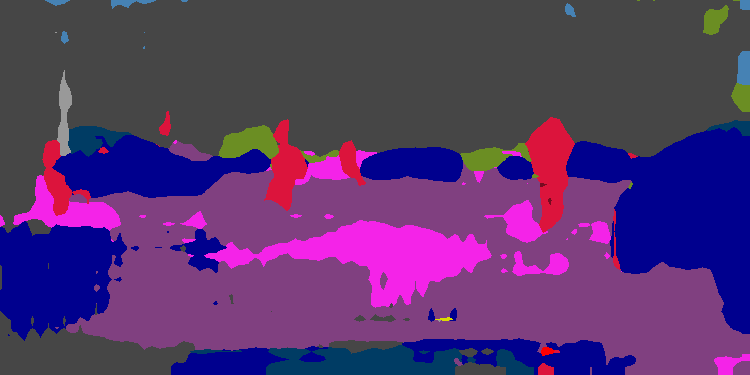} &  
  \includegraphics[width=\imgsize]{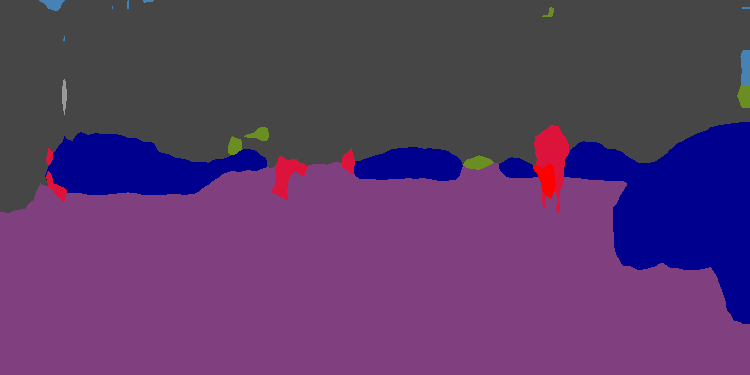} & 
  \includegraphics[width=\imgsize]{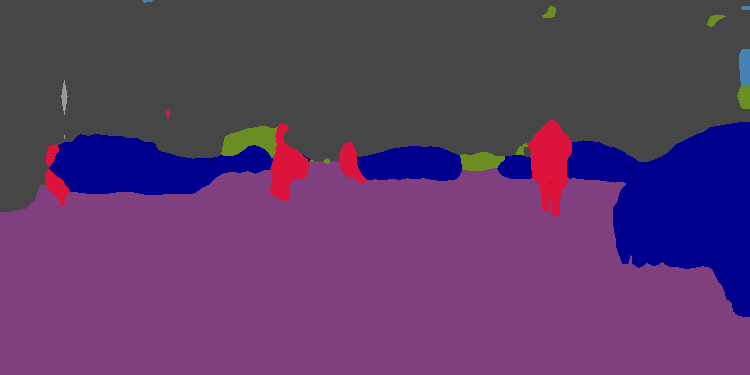} & 
  \includegraphics[width=\imgsize]{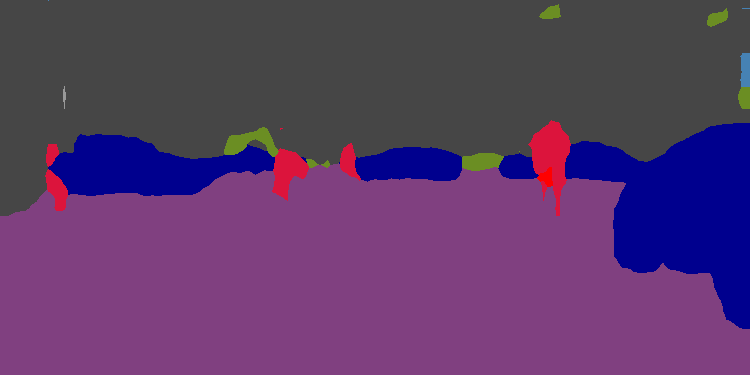} \\
 
  && & \includegraphics[width=\imgsize]{images/DA_GTA-Cityscapes/rgb/frankfurt_000001_055172_leftImg8bit.png} &
  \includegraphics[width=\imgsize]{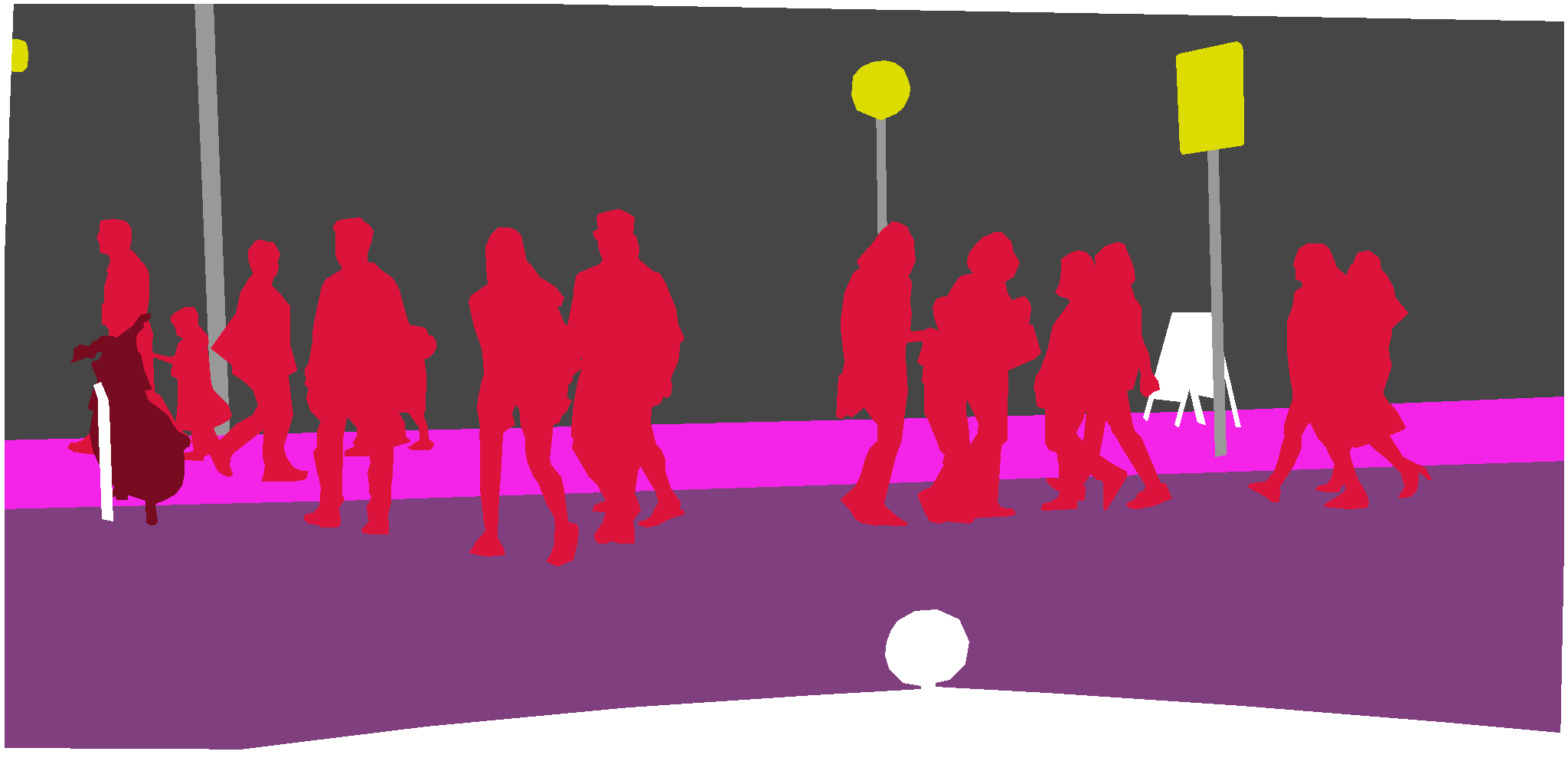} & 
  \includegraphics[width=\imgsize]{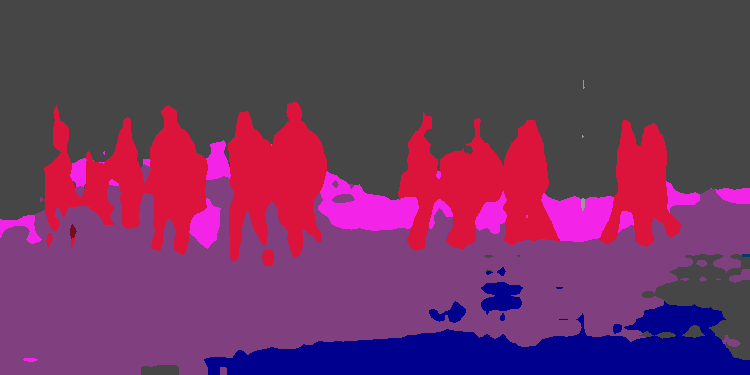} & 
  \includegraphics[width=\imgsize]{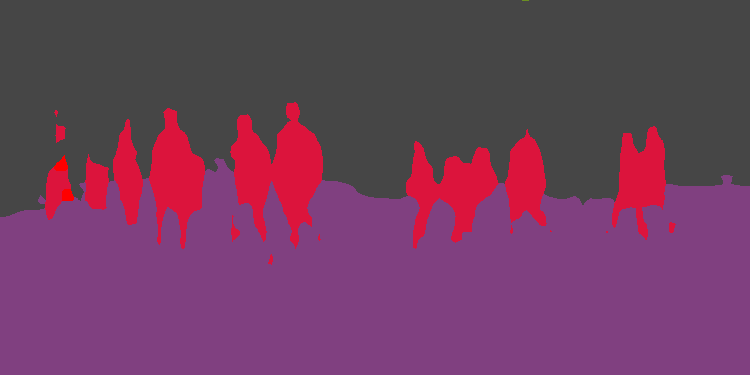} & 
  \includegraphics[width=\imgsize]{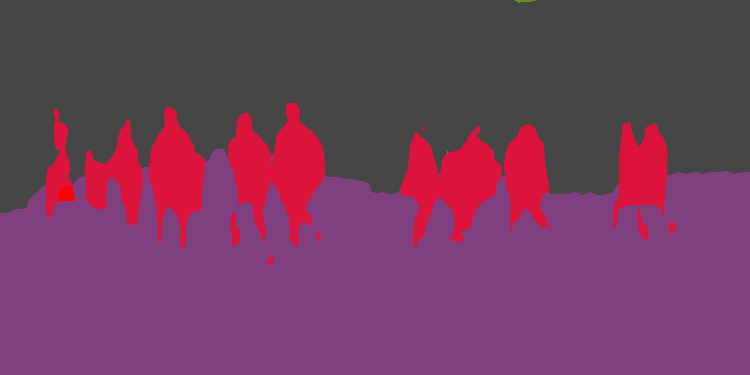} & 
  \includegraphics[width=\imgsize]{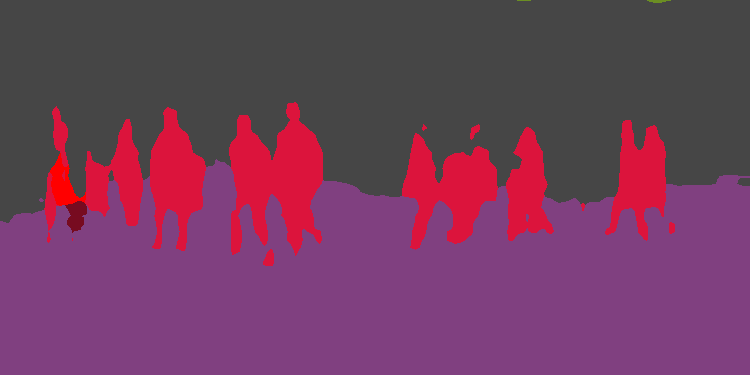} \\
  
  && & \includegraphics[width=\imgsize]{images/DA_GTA-Cityscapes/rgb/frankfurt_000000_006589_leftImg8bit.png} &
  \includegraphics[width=\imgsize]{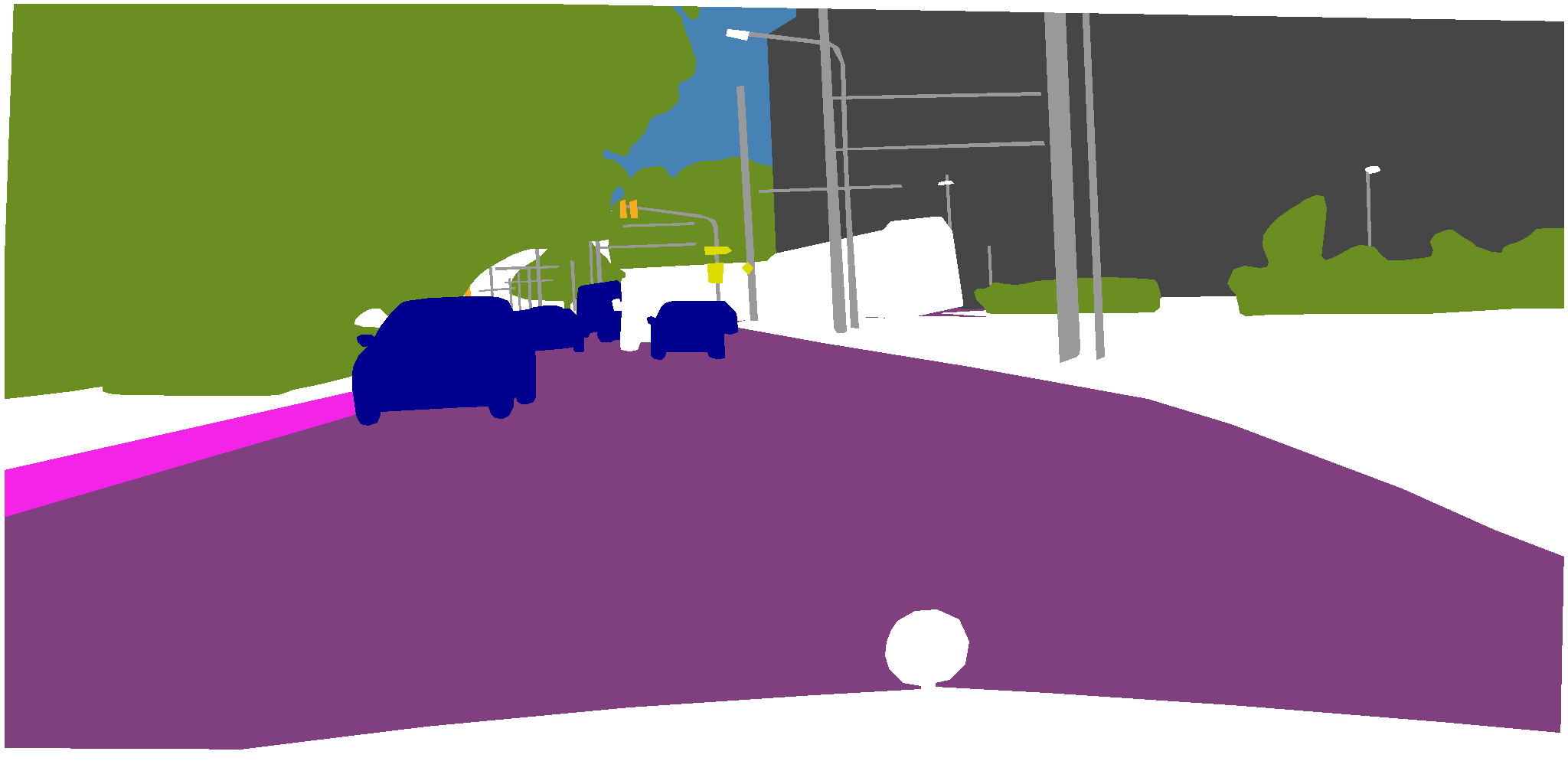} & 
  \includegraphics[width=\imgsize]{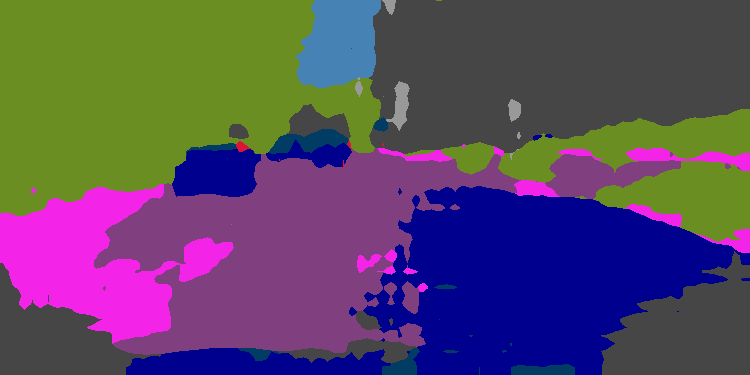} & 
  \includegraphics[width=\imgsize]{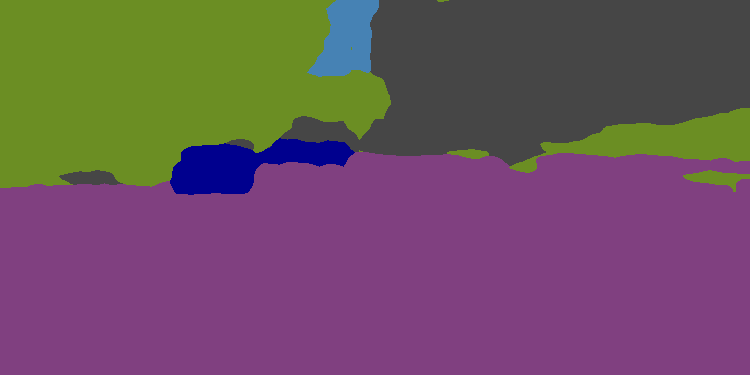} & 
  \includegraphics[width=\imgsize]{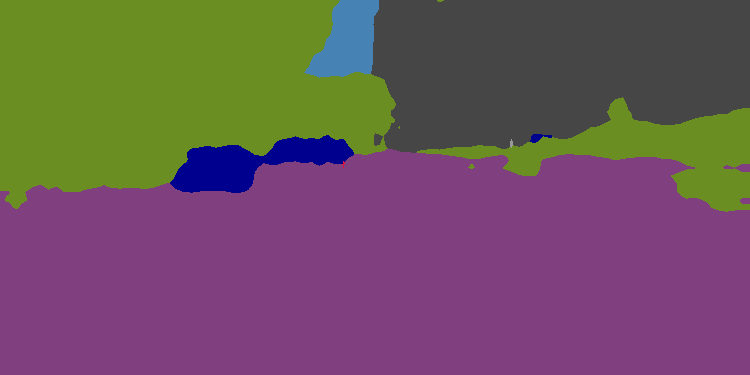} & 
  \includegraphics[width=\imgsize]{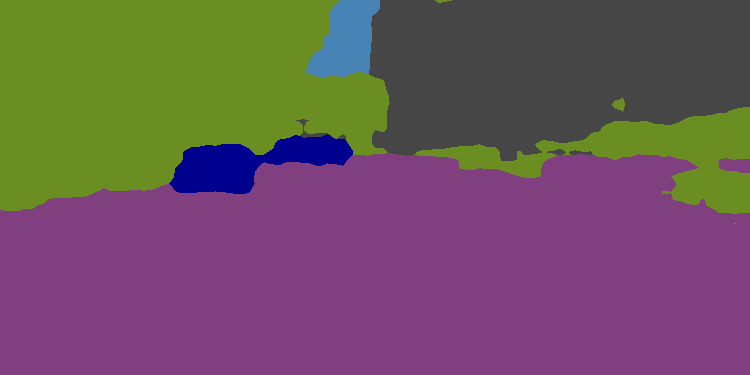} \\
  \cline{2-3}
  
\end{tabular}
\end{subfigure}

\vspace{0.4cm}
\centering
\begin{subfigure}[htbp]{2\textwidth}
\begin{tabular}{c|c|c|cccccc}

\cline{2-3}
    \multirow{6}{*}{\vspace{-27ex}b)} & \multirow{6}{*}{\rotatebox{90}{\hspace{-20ex}To Mapillary}} & \multirow{3}{*}{\rotatebox{90}{\hspace{-7ex}From GTA5}} &\includegraphics[width=\imgsize]{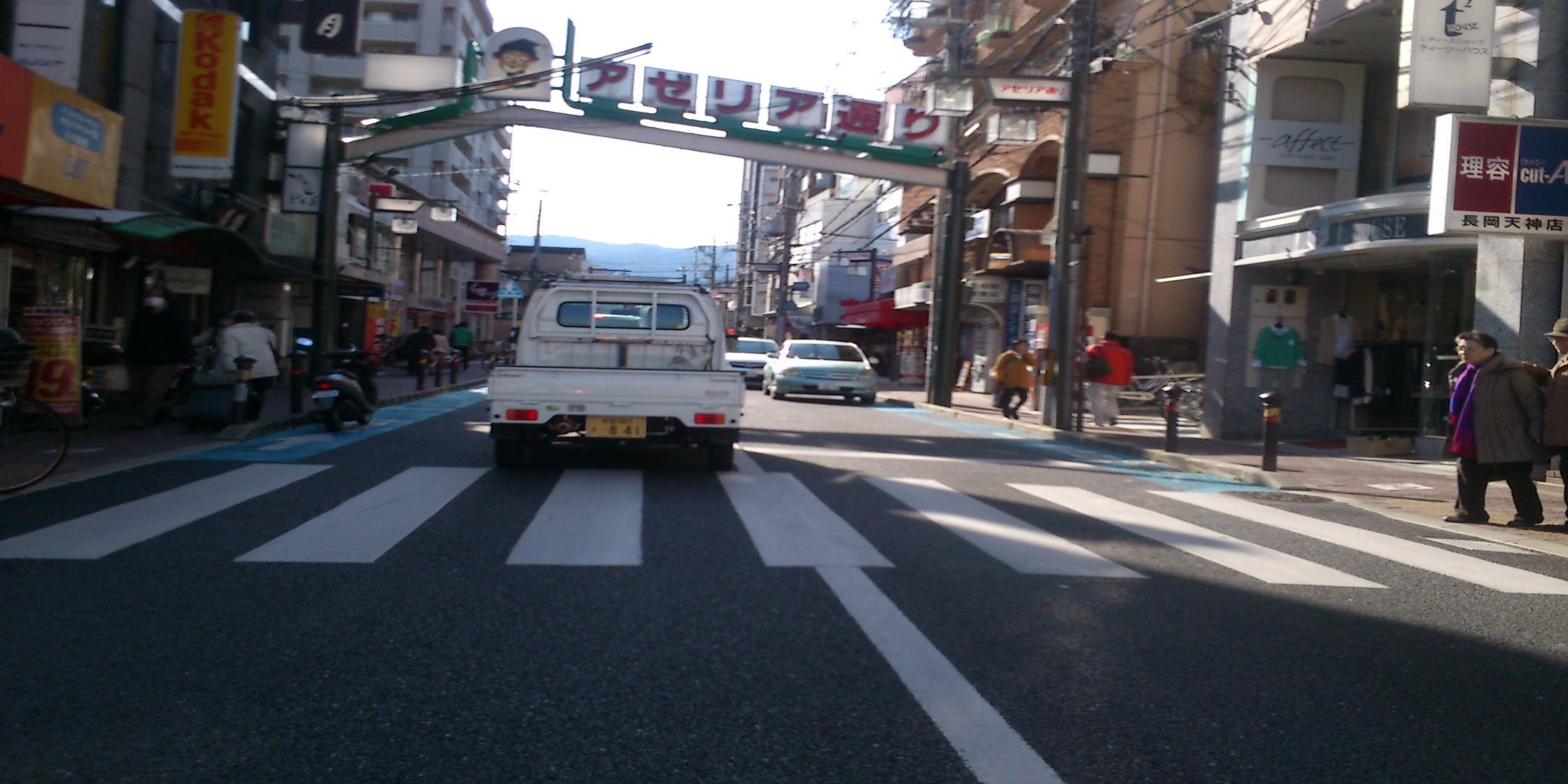} &
   \includegraphics[width=\imgsize]{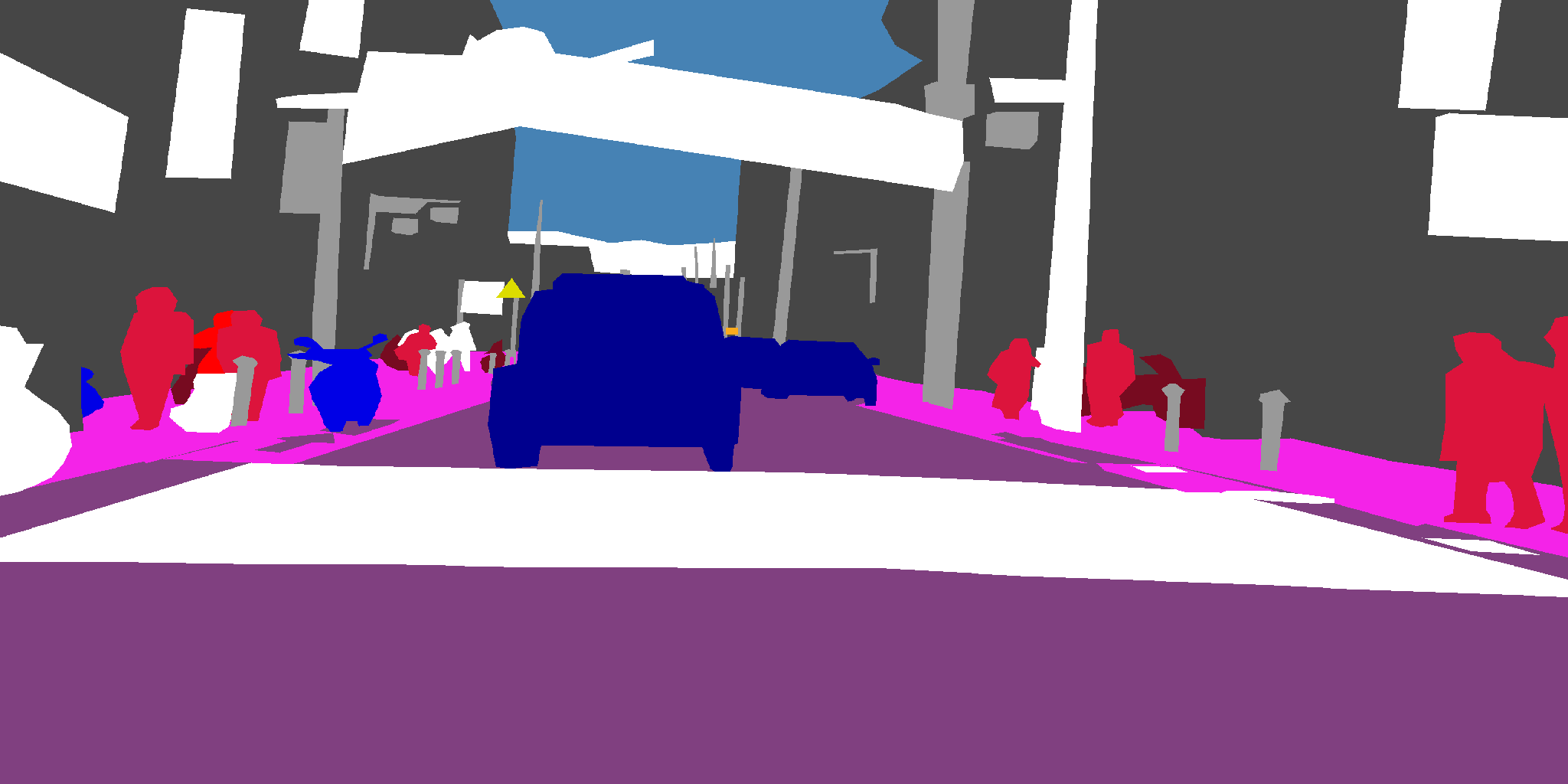} & 
   \includegraphics[width=\imgsize]{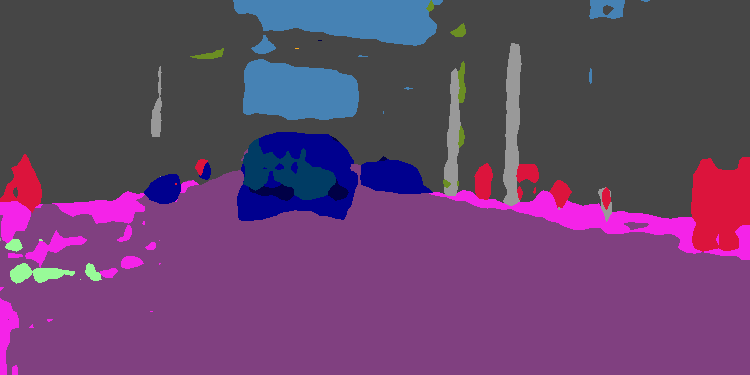} & 
   \includegraphics[width=\imgsize]{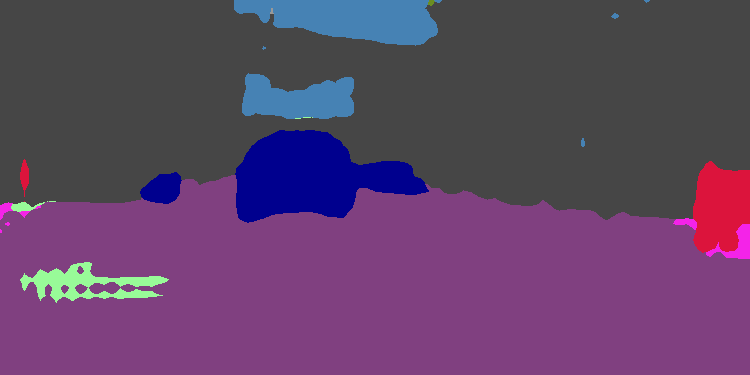} & 
   \includegraphics[width=\imgsize]{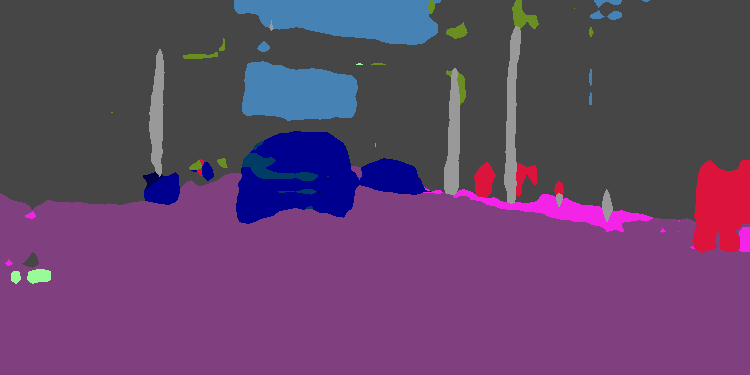} & 
   \includegraphics[width=\imgsize]{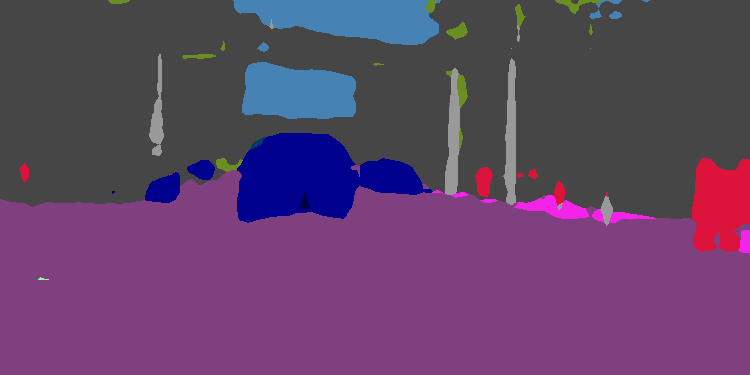} \\
  
  && & \includegraphics[width=\imgsize]{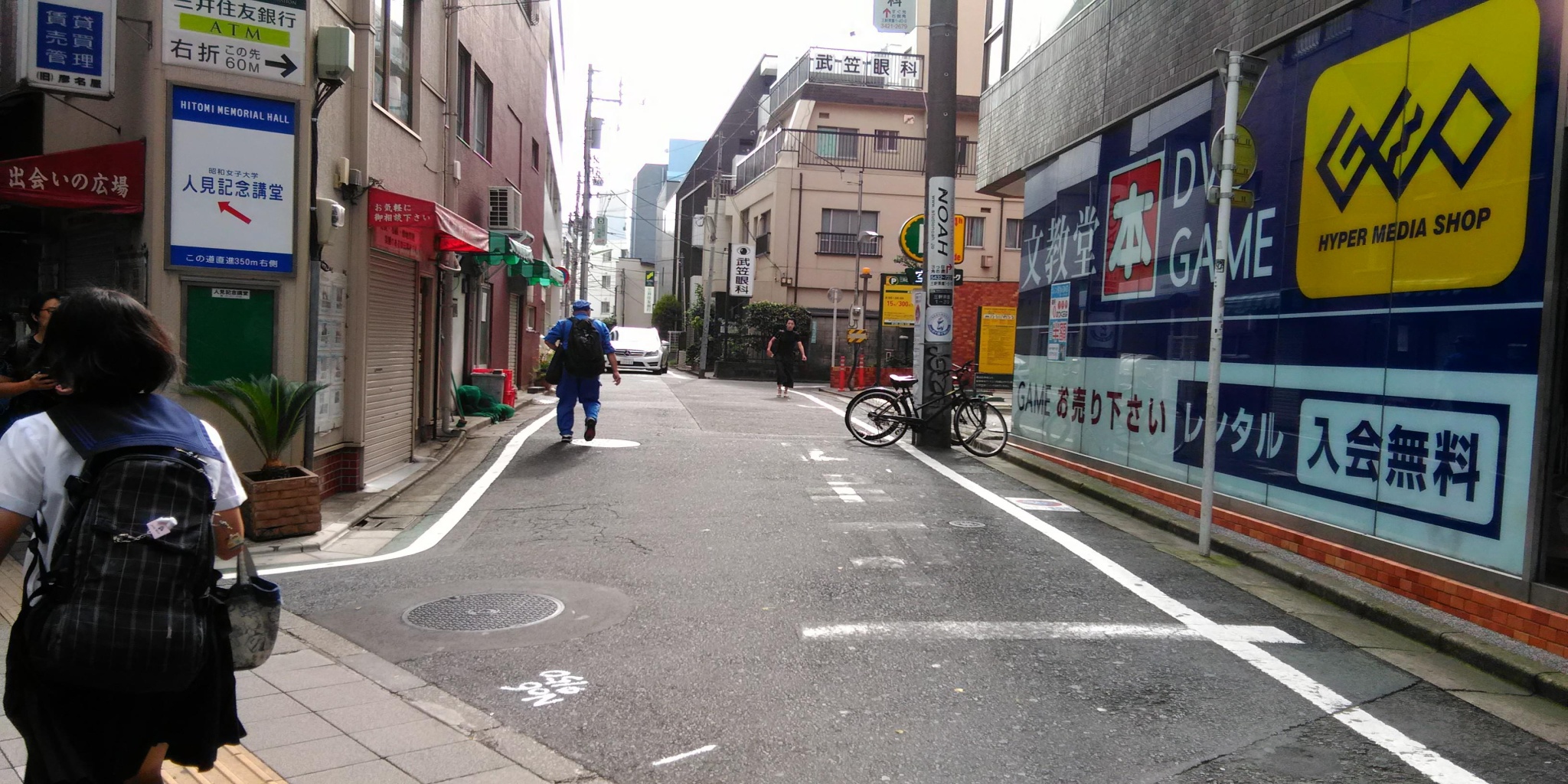} &
   \includegraphics[width=\imgsize]{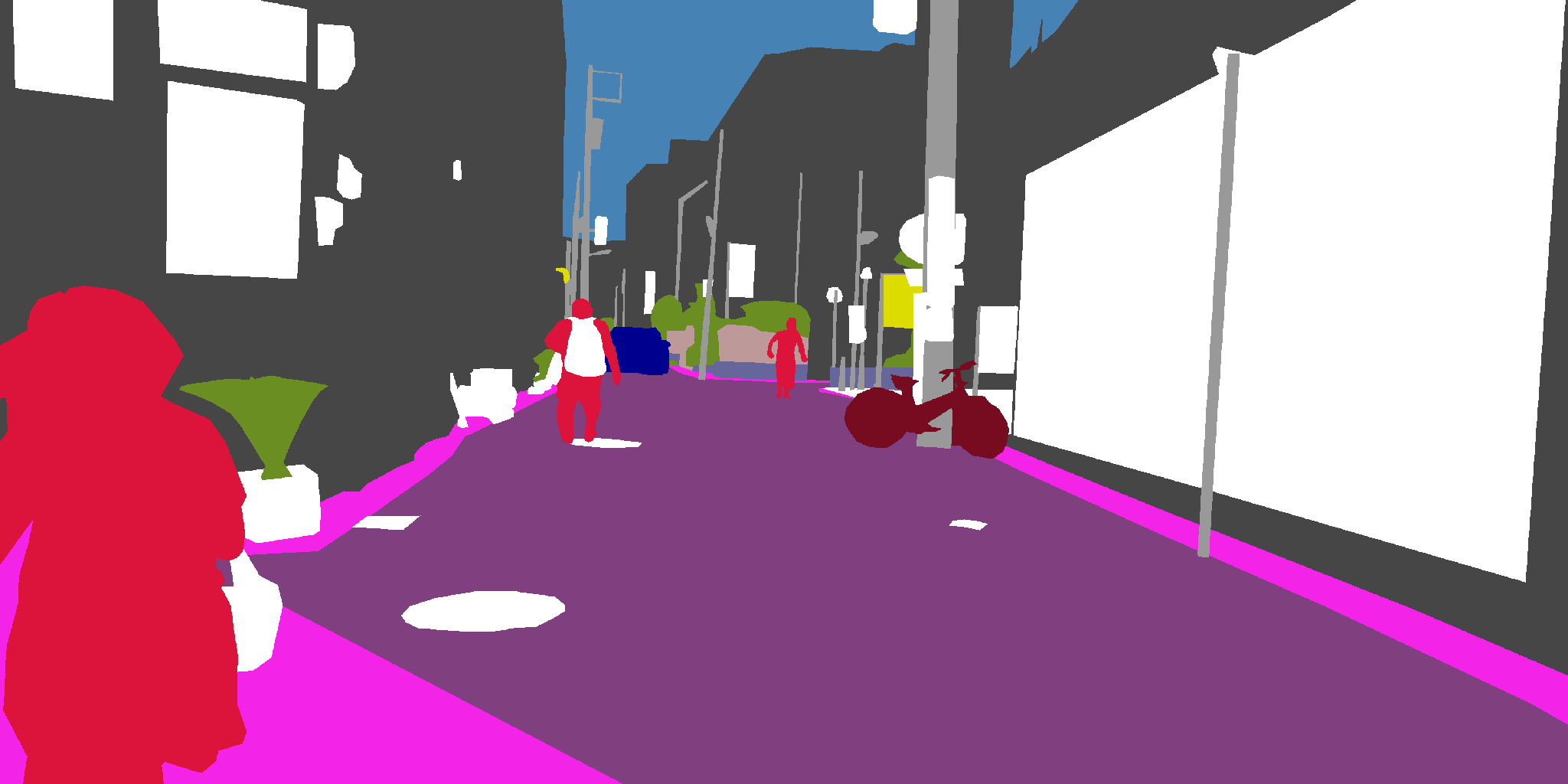} & 
   \includegraphics[width=\imgsize]{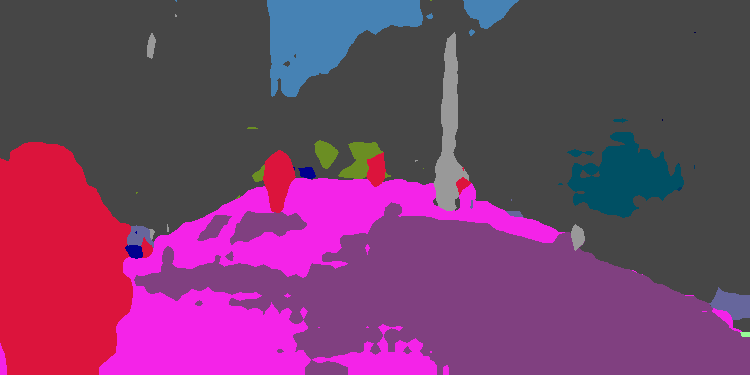} & 
   \includegraphics[width=\imgsize]{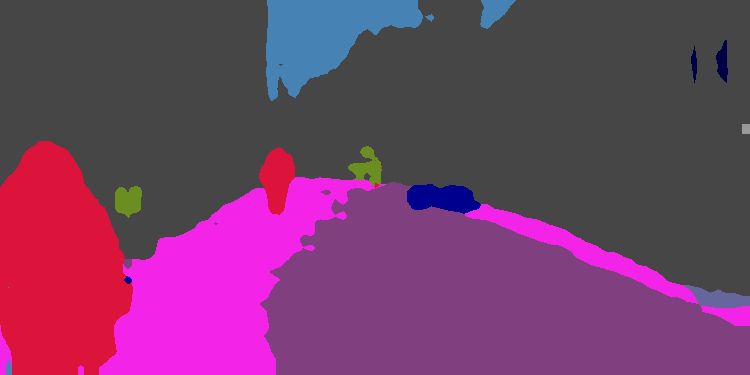} & 
   \includegraphics[width=\imgsize]{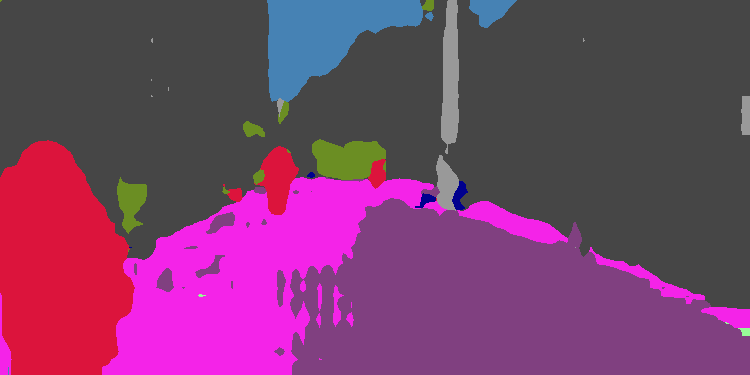} & 
   \includegraphics[width=\imgsize]{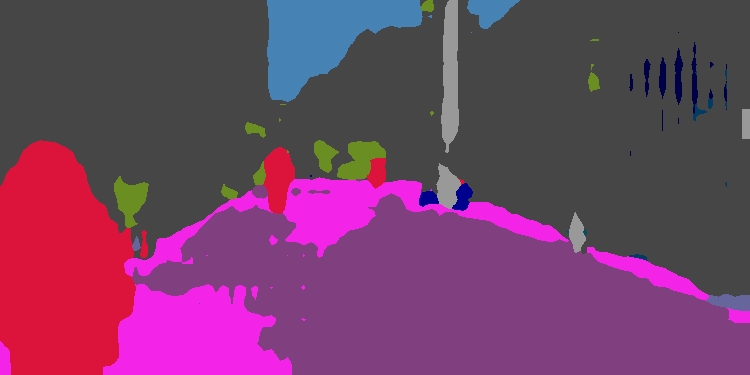} \\

   && & \includegraphics[width=\imgsize]{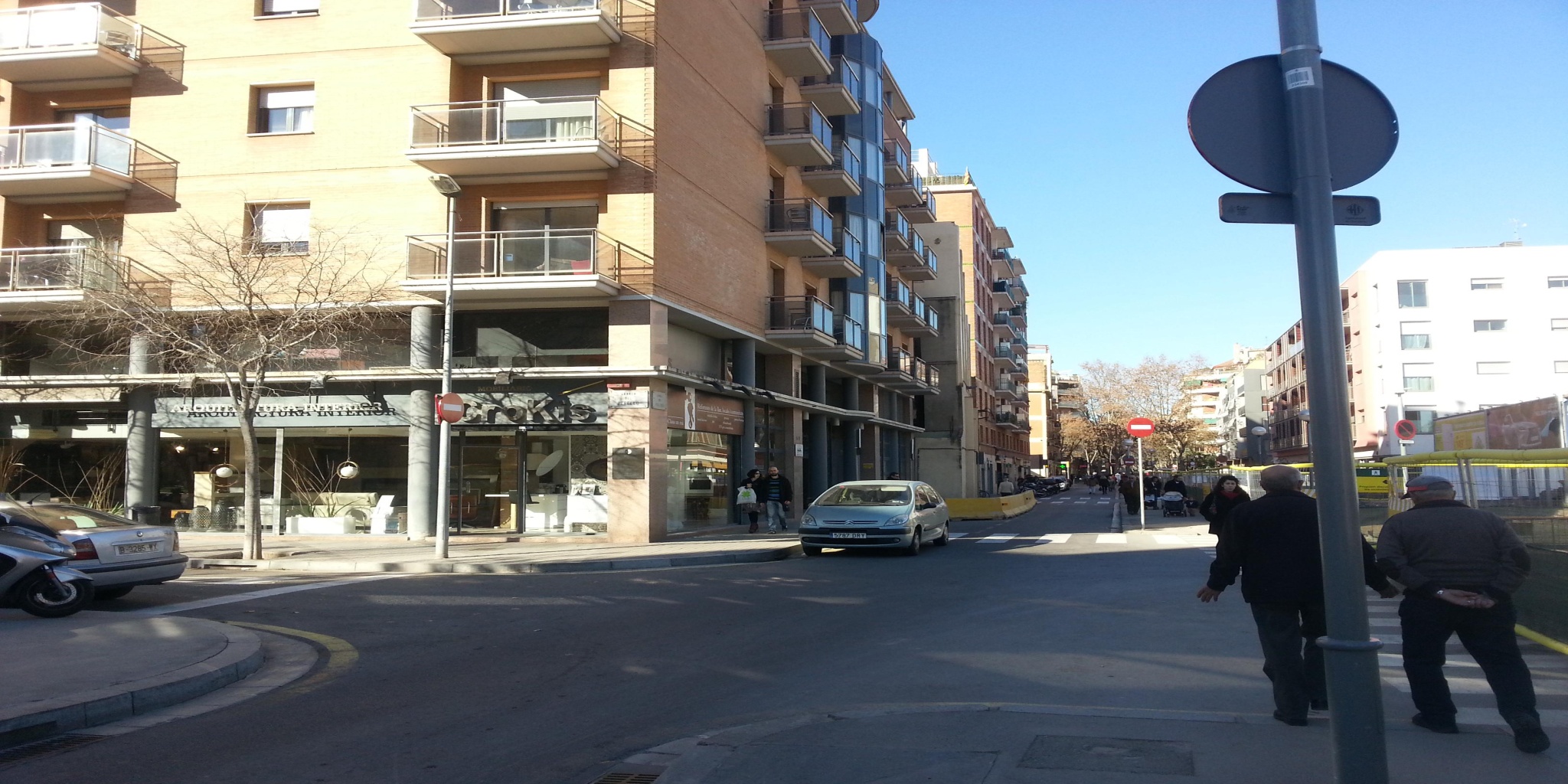} &
   \includegraphics[width=\imgsize]{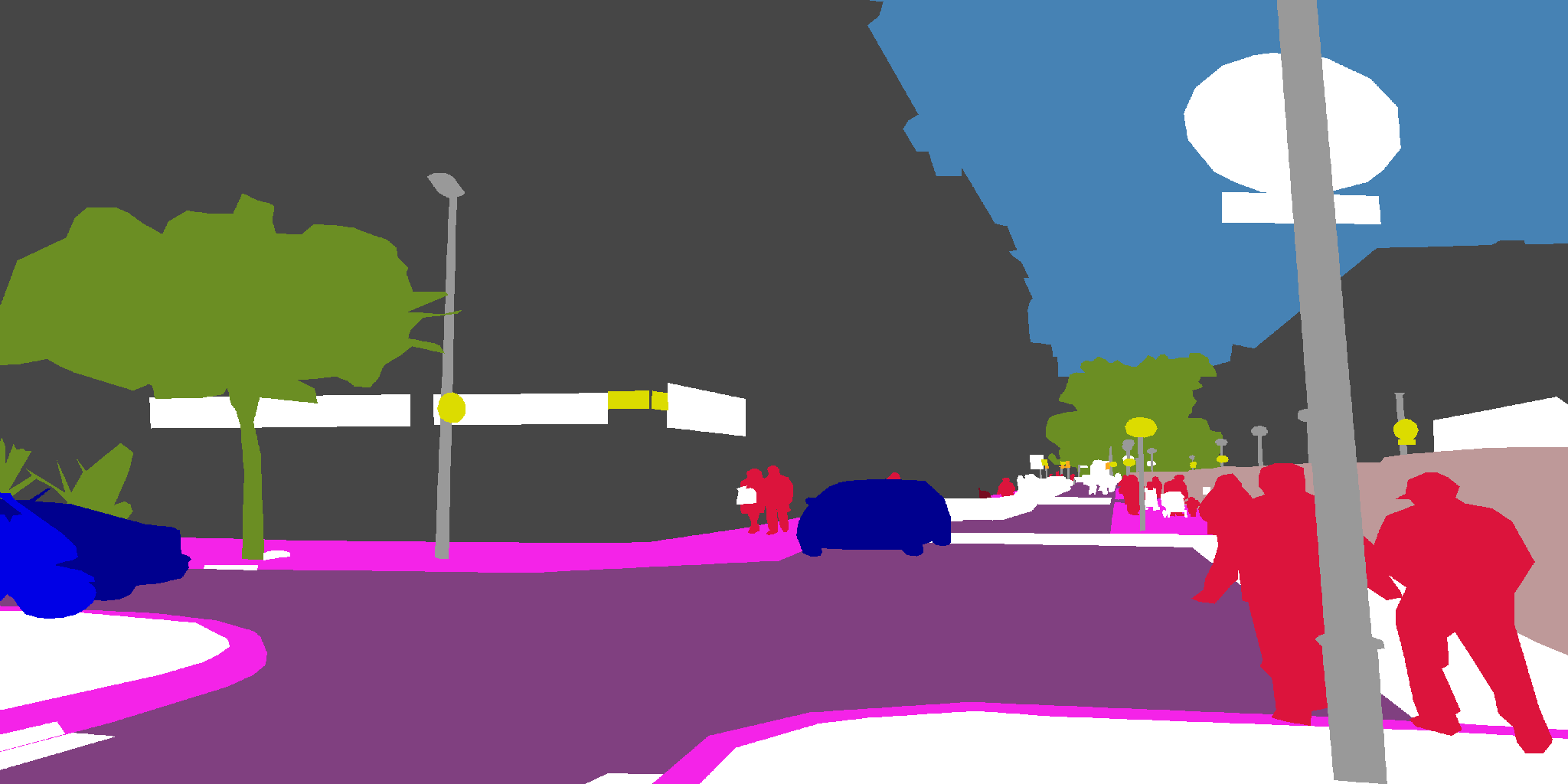} & 
   \includegraphics[width=\imgsize]{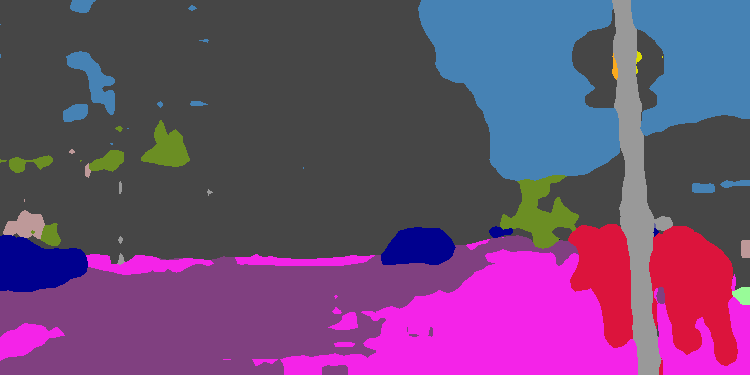} & 
   \includegraphics[width=\imgsize]{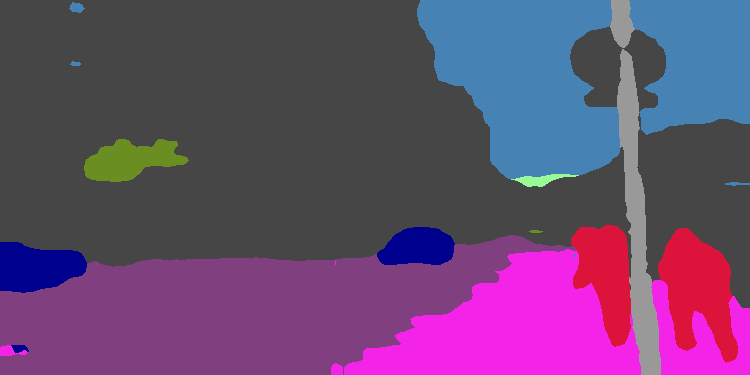} & 
   \includegraphics[width=\imgsize]{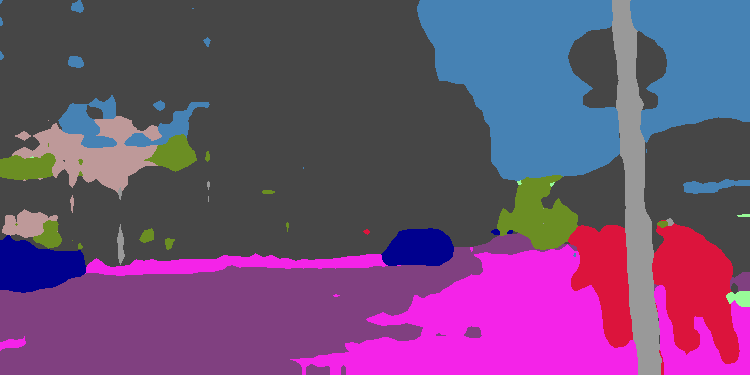} & 
   \includegraphics[width=\imgsize]{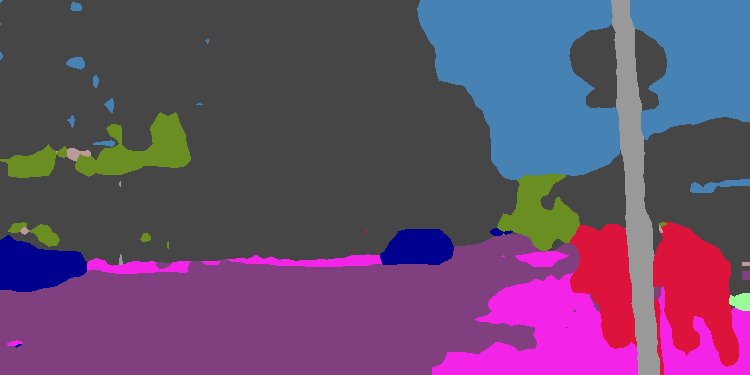} \\\cline{3-3}
  
   && \multirow{3}{*}{\rotatebox{90}{\hspace{+3ex}From SYNTHIA}} & 
   \includegraphics[width=\imgsize]{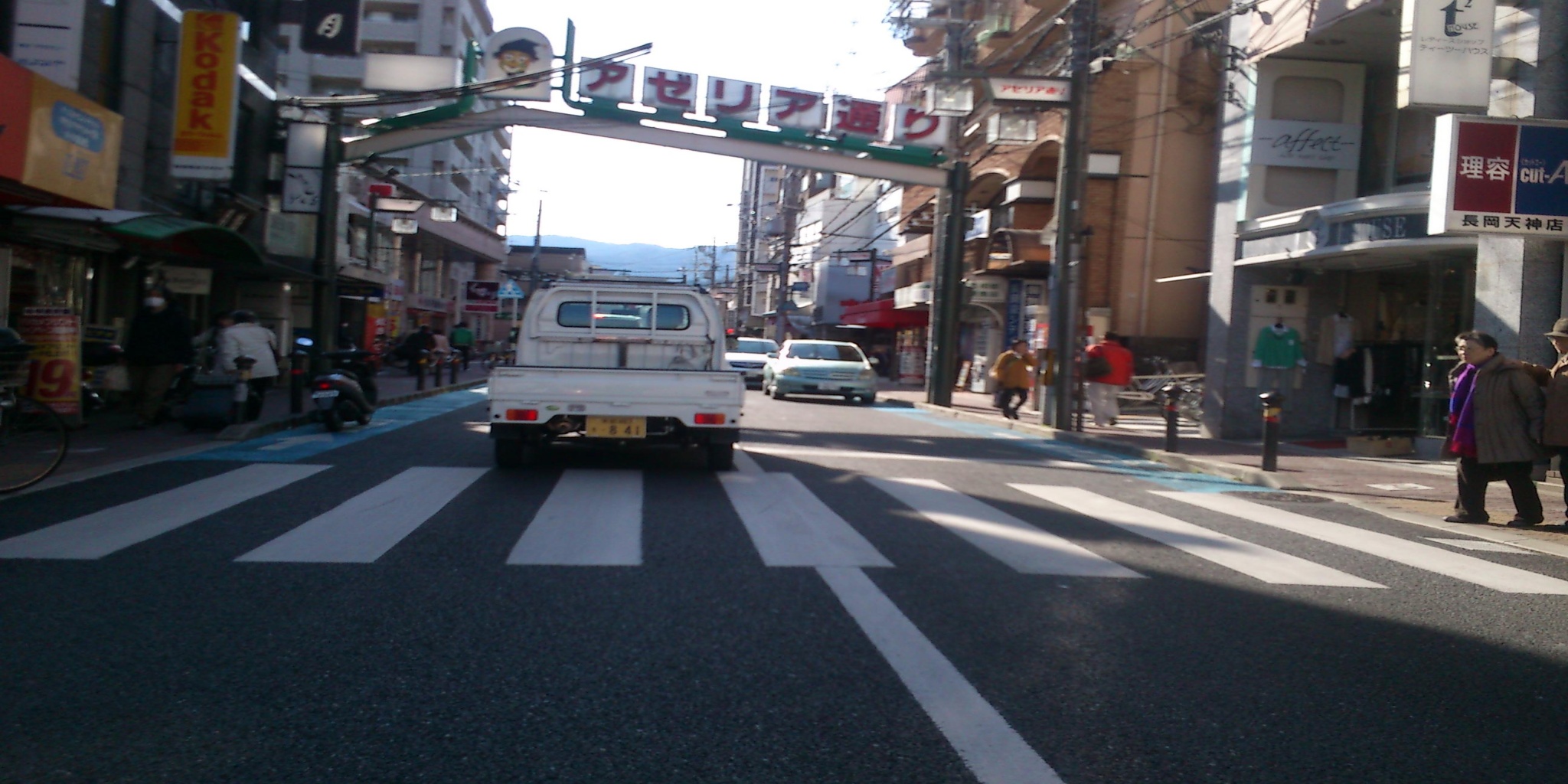} &
  \includegraphics[width=\imgsize]{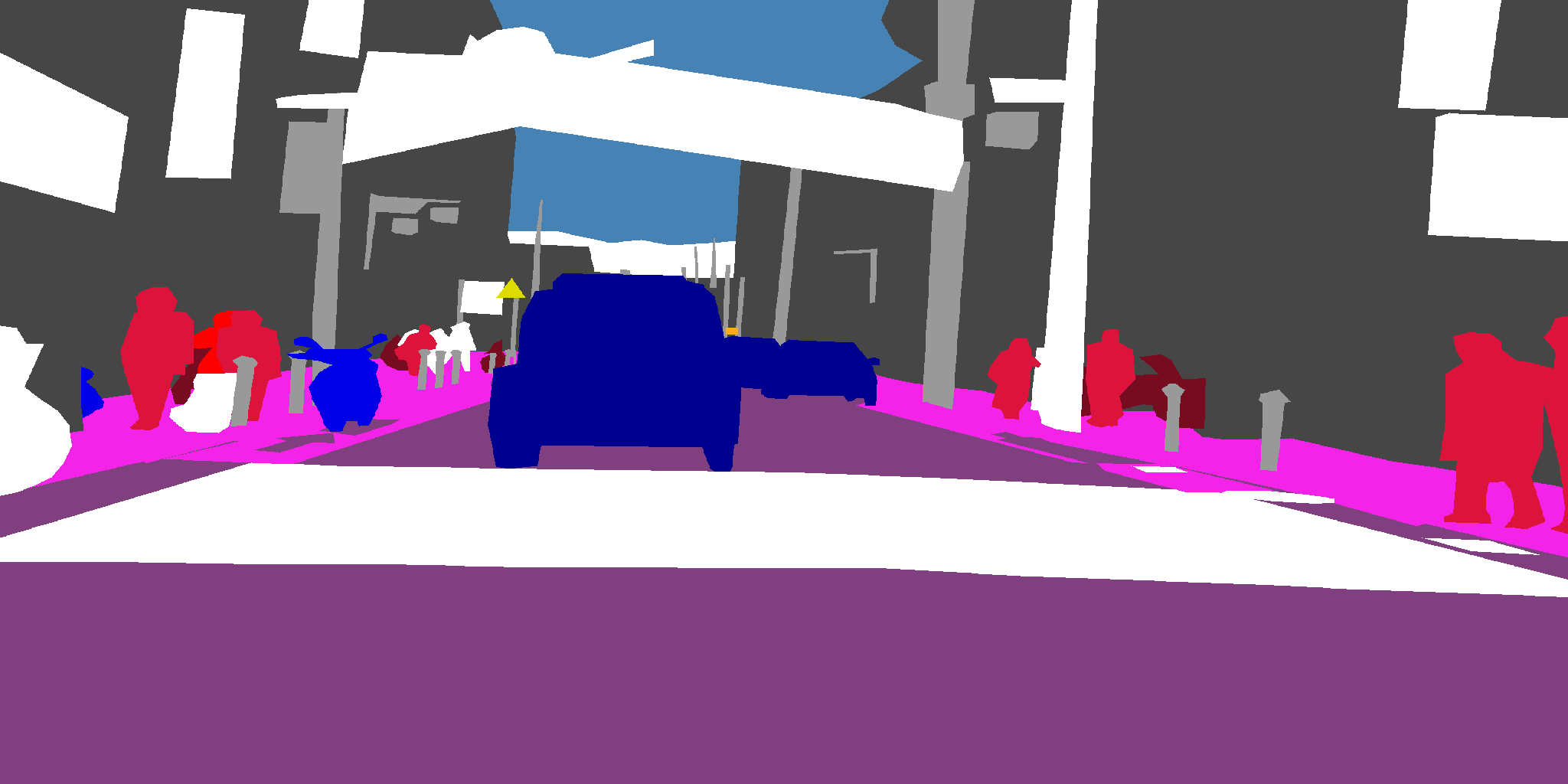} & 
  \includegraphics[width=\imgsize]{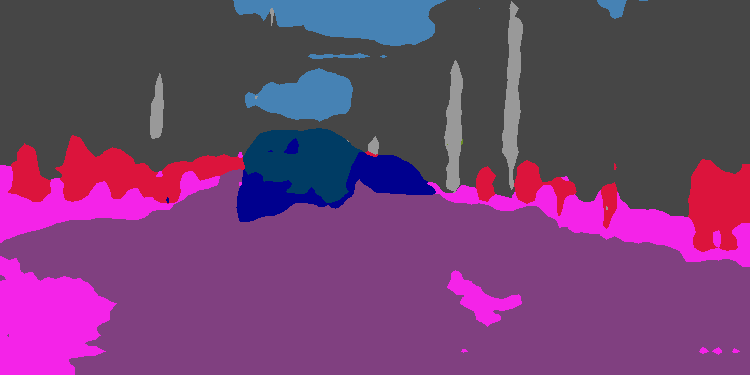} &  
  \includegraphics[width=\imgsize]{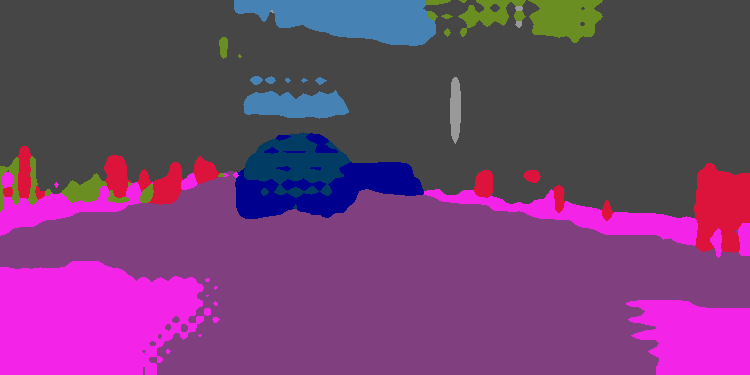} & 
  \includegraphics[width=\imgsize]{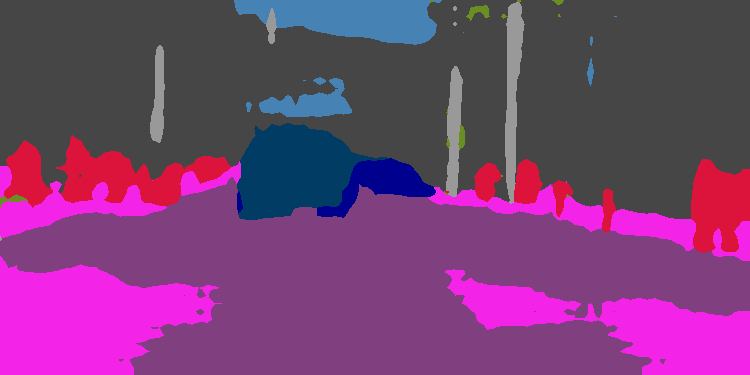} & 
  \includegraphics[width=\imgsize]{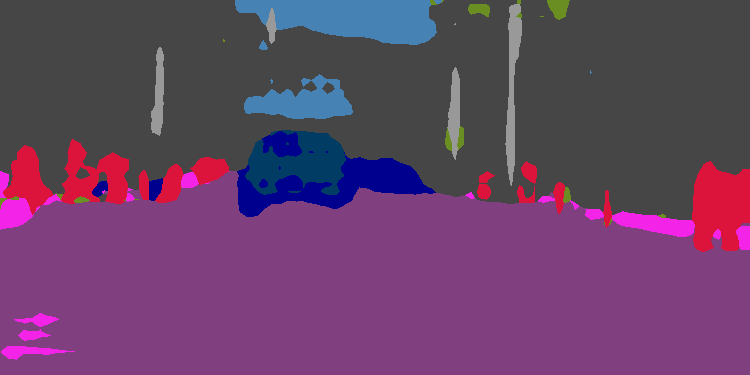} \\

 && & \includegraphics[width=\imgsize]{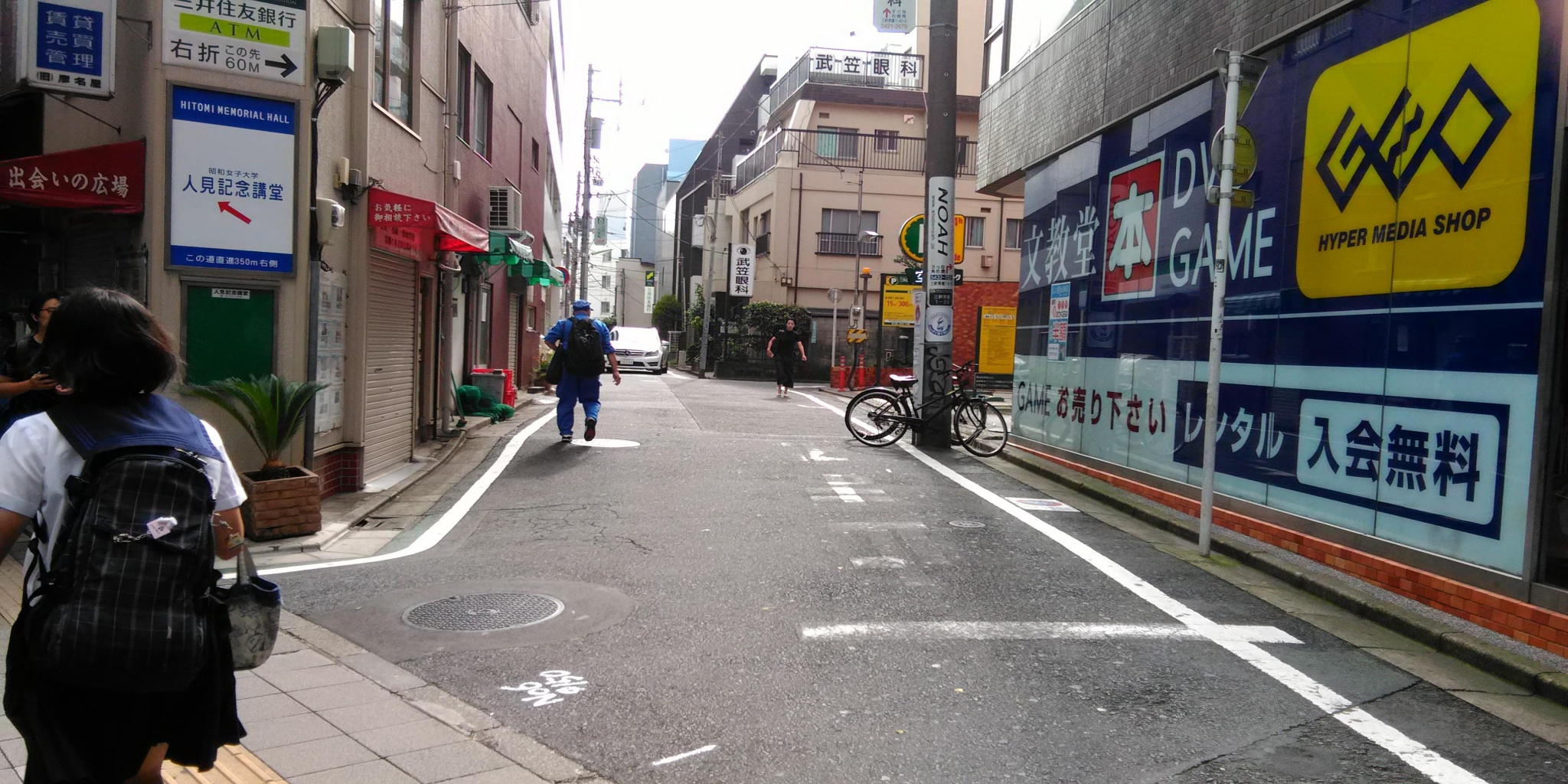} &
  \includegraphics[width=\imgsize]{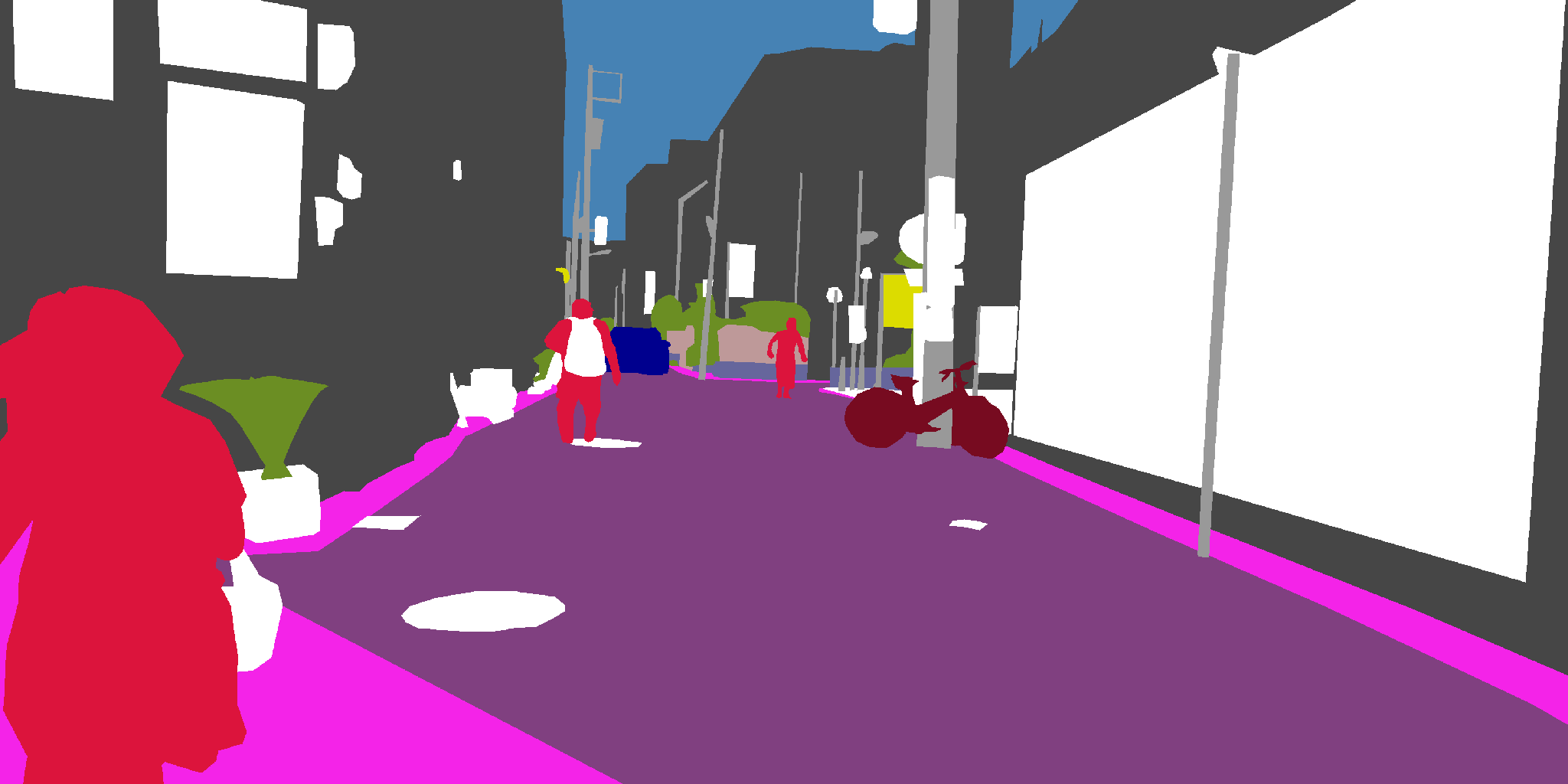} & 
  \includegraphics[width=\imgsize]{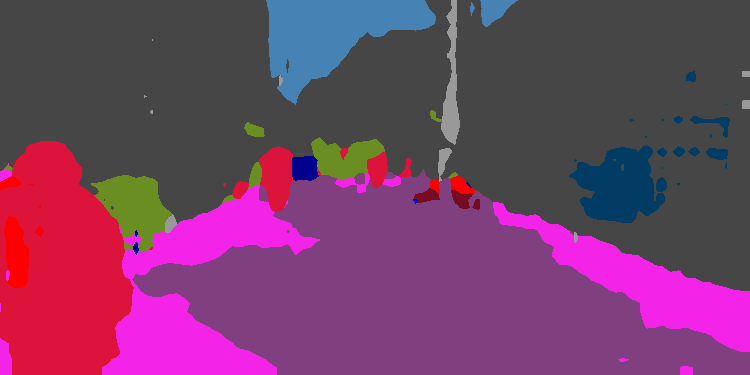} & 
  \includegraphics[width=\imgsize]{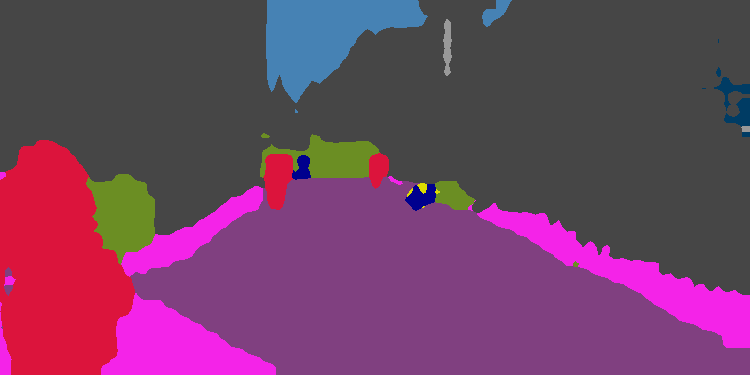} & 
  \includegraphics[width=\imgsize]{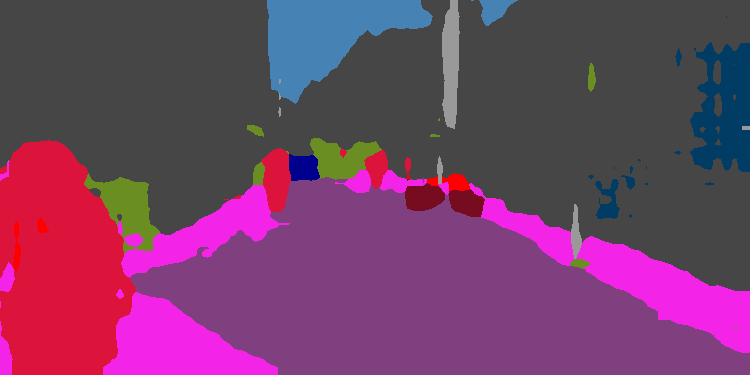} & 
  \includegraphics[width=\imgsize]{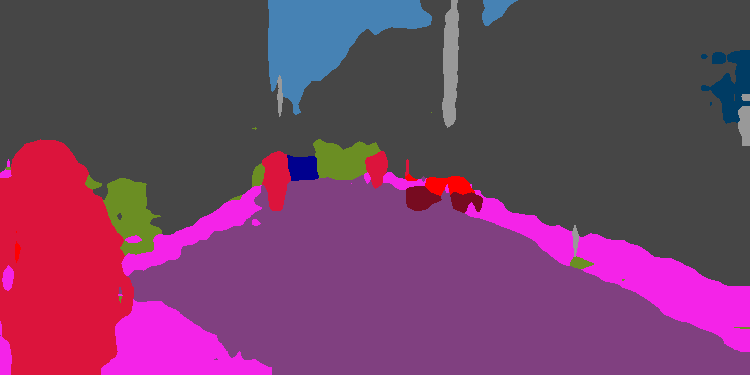} \\

  & && \includegraphics[width=\imgsize]{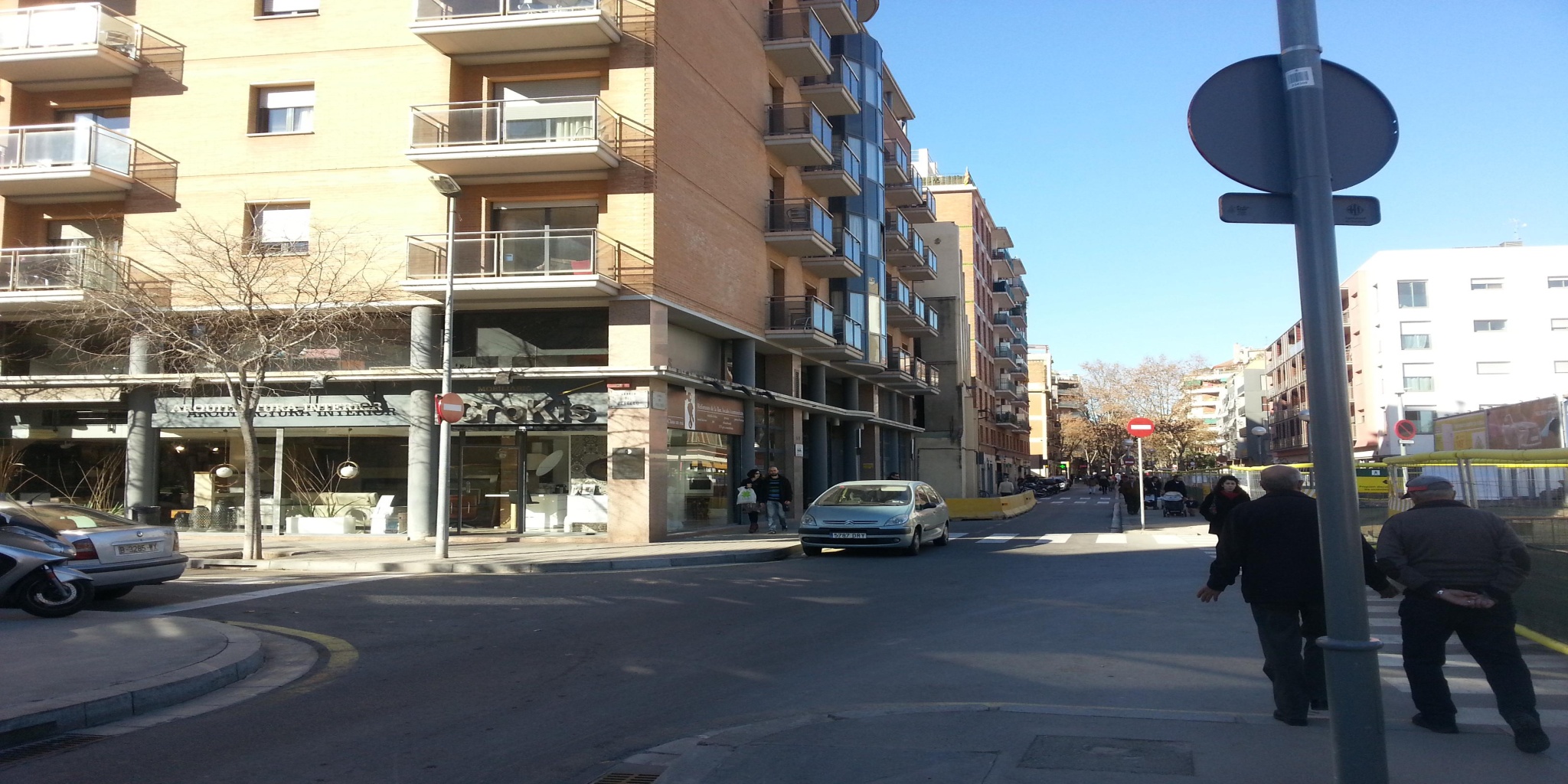} &
  \includegraphics[width=\imgsize]{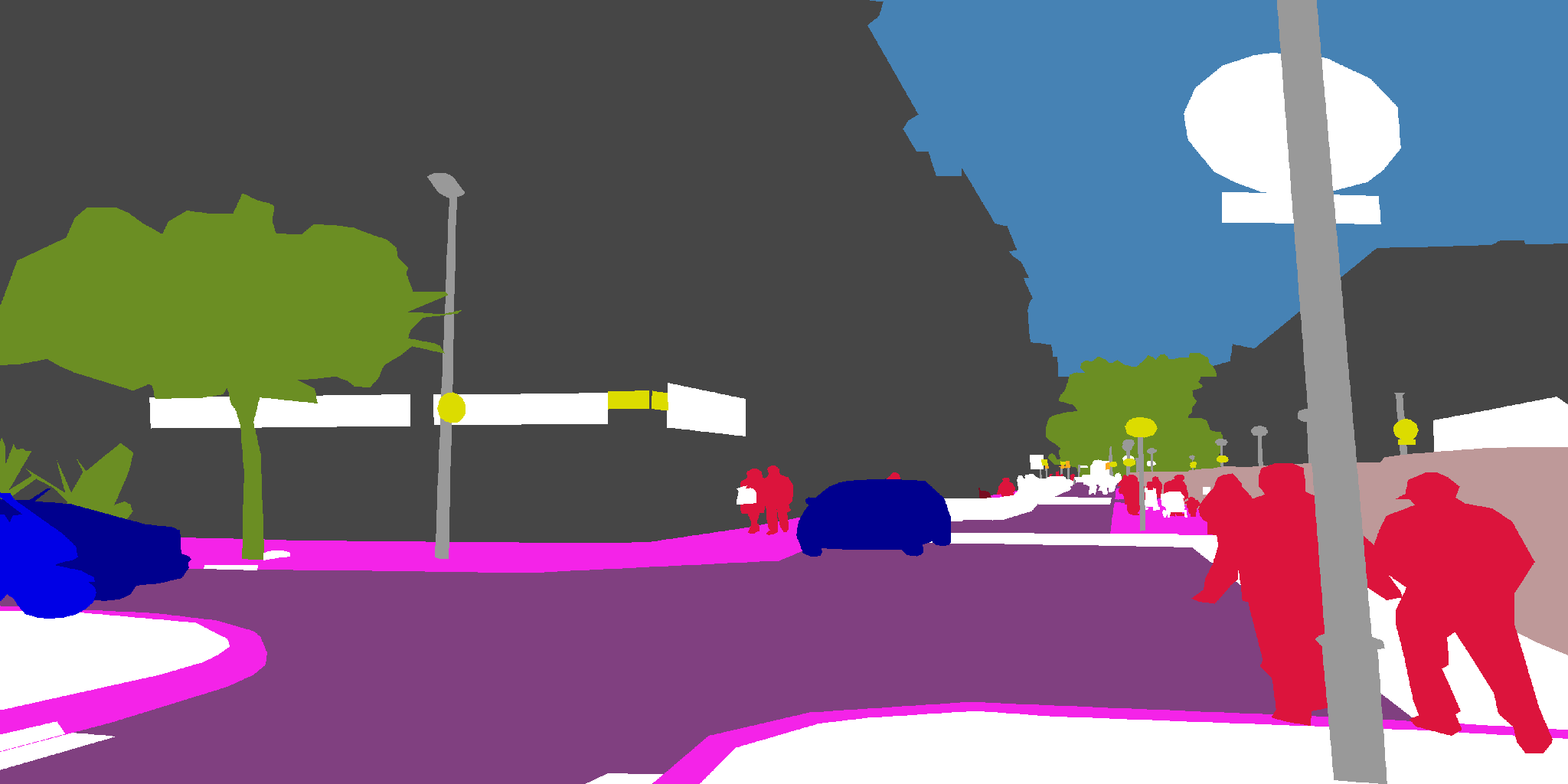} & 
  \includegraphics[width=\imgsize]{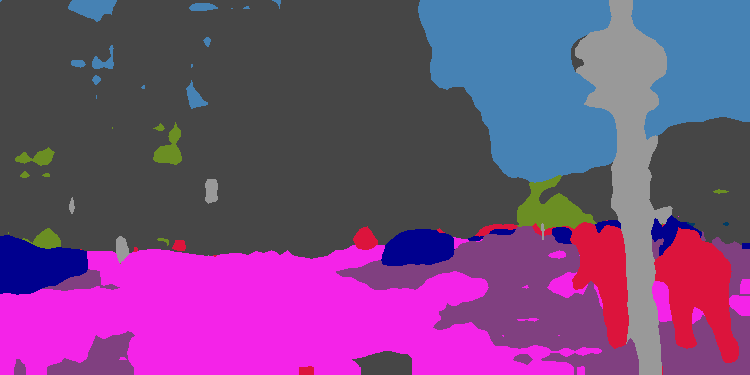} & 
  \includegraphics[width=\imgsize]{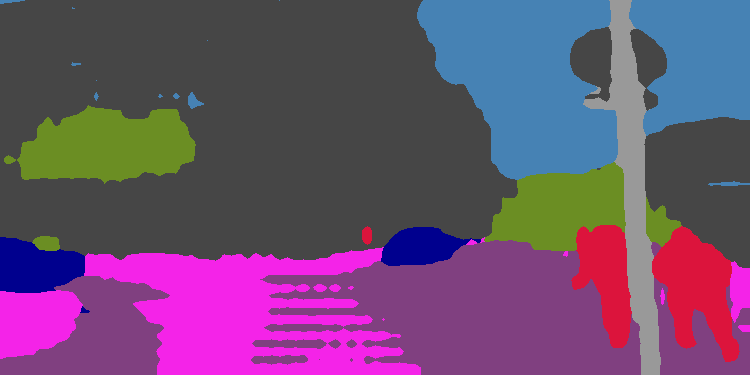} & 
  \includegraphics[width=\imgsize]{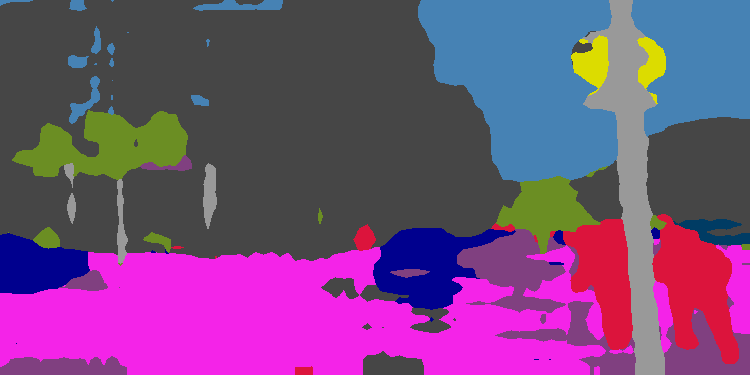} & 
  \includegraphics[width=\imgsize]{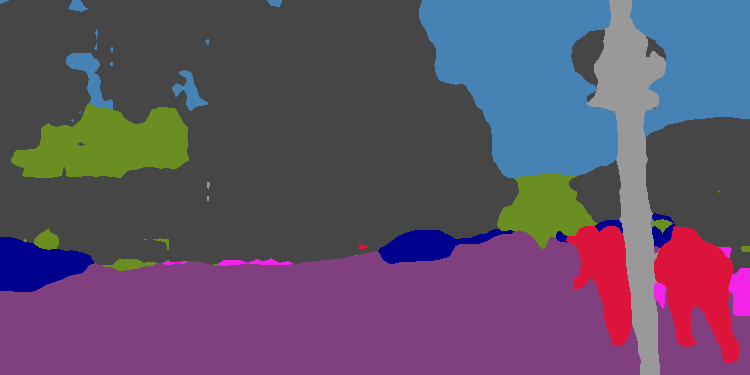} \\
 \cline{2-3}
 \multicolumn{3}{c}{} & Image & Annotation & Supervised ($\mathcal{L}_{G,1}$) & Hung et al. \cite{hung2018} & Biasetton et al. \cite{biasetton2019} & Ours ($\mathcal{L}_{full}$)
\end{tabular}
\end{subfigure}
\caption{Semantic segmentation of some sample scenes extracted from the Cityscapes (a) and Mapillary (b) validation sets. The first group of six rows is related to the Cityscapes dataset, the last six to the Mapillary dataset. For each group, the first three rows are related to the experiments in which the GTA5 dataset is used as source. The last three rows are related to the case in which the SYNTHIA dataset is used as source (\textit{best viewed in colors}).}
\label{fig:qual_res}
\end{figure*}

\reviewtxt{Adapting from SYNTHIA,} the task is even more challenging w.r.t. the GTA5 case since the computer generated graphics are less realistic. \reviewtxt{By training the network $G$ in a supervised way on SYNTHIA} and performing inference on the real world  Cityscapes dataset, a mIoU of $25.4\%$ can be obtained (see Table \ref{tab:SYNTHIA}). This value is smaller than the mIoU of $27.9\%$ obtained by training $G$ on the GTA5 dataset. 
The performance gap confirms that the GTA5 dataset has a smaller domain shift with respect to real world data, when compared with the SYNTHIA dataset. 
By exploiting the proposed approach an accuracy of $31.3\%$ can be obtained. The improvement is very similar to the one obtained using GTA5 as source dataset, proving that the approach is able to generalize to different datasets. In this case, there is a larger variability among different classes, however notice the very large improvement on \textit{road} and \textit{building} classes.
The previous version of the method \cite{biasetton2019} has an accuracy of $30.2\%$

Furthermore, our framework outperforms the compared state-of-the-art approaches. The method of Hung et al. \cite{hung2018}, \reviewtxt{that exploits the} same generator architecture of our approach, obtains a mIoU equal to $29.4\%$. The approach of \cite{zhang2017} has an even lower mIoU of $29.0\%$.  The method of \cite{hoffman2016} is  the less performing approach and in this comparison it is even less accurate than our  synthetic supervised trained network, however it employs a different segmentation network.

The fourth, fifth and sixth row of Figure \ref{fig:qual_res}a shows the output on 
the same sample scenes discussed above when the SYNTHIA dataset is used as source instead of GTA5. 
\reviewtxt{The first thing that stands out is that }
 by training on the SYNTHIA dataset some very common classes as \textit{sidewalk} and \textit{road} are highly corrupted.
This is caused by the not very realistic textures used \reviewtxt{for such classes} in the SYNTHIA dataset. 
Furthermore, while the positioning of the camera in the Cityscapes dataset is always fixed and mounted on-board inside the car, in SYNTHIA the camera can be placed in different positions. For example, the pictures can be captured from inside the car, \reviewtxt{from the top} or from the side of the road.

The approach of Hung et al. \cite{hung2018} is able to correctly recognize the class \textit{road}, correcting the noise present in the synthetic supervised training. \reviewtxt{However, as mentioned earlier,} it suffers on small classes where it tends to lose small objects and to produce \reviewtxt{imprecise shapes.}
The method of \cite{biasetton2019} and the proposed one have slightly better \reviewtxt{performances: the} last two columns of Figure \ref{fig:qual_res}a show how the unsupervised adaptation and the self-teaching \reviewtxt{component allow to avoid all the artifacts on the \textit{road} surface. The segmentation network now captures the real nature of this class in the Cityscapes dataset.} At the same time, 
our method is able to locate a bit more precisely small classes as \textit{person} and \textit{vegetation}. However, in this setting the difference between the old and new version of the proposed method is limited.

\begin{table*}[htbp]
\setlength{\tabcolsep}{1.6pt}
\centering
\begin{tabular}{|c|ccccccccccccccccccc|c|}
\hline
\textbf{GTA5} $\rightarrow$ \textbf{Mapillary}  & \rotatebox{90}{road} &  \rotatebox{90}{sidewalk} &  \rotatebox{90}{building} &  \rotatebox{90}{wall} &\rotatebox{90}{fence} & \rotatebox{90}{pole} & \rotatebox{90}{t light} 
  &\rotatebox{90}{t sign} & \rotatebox{90}{veg} & \rotatebox{90}{terrain} & \rotatebox{90}{sky} & \rotatebox{90}{person}& \rotatebox{90}{rider} & \rotatebox{90}{car} 
  & \rotatebox{90}{truck} & \rotatebox{90}{bus} & \rotatebox{90}{train} &  \rotatebox{90}{mbike} & \rotatebox{90}{bike} & \rotatebox{90}{mean} \\
 \hline
Supervised ($\mathcal{L}_{G,1}$ only) & 66.5 & 24.4 & 46.1 & 17.9 & 21.6 & 24.8 & \textbf{11.8} & \textbf{5.9} & 70.7 & 25.6 & 66.1 & 57.3 & 10.2 & 79.7 & 37.3 & 39.8 & 4.6 & 10.1 & \textbf{1.7} & 32.7 \\
Ours (full) & \textbf{79.9} & 28.0 & \textbf{73.4} & \textbf{23.0} & \textbf{29.5} & 20.9 & 1.1 & 0.0 & \textbf{79.5} & \textbf{39.6} & 95.0 & 57.6 & 9.0 & \textbf{80.6} & \textbf{41.5} & 40.1 & \textbf{7.4} & \textbf{24.8} & 0.1 & \textbf{38.5} \\ \hline
Hung et al. \cite{hung2018} & 78.2 & \textbf{29.7} & 68.7 & 10.0 & 6.7 & 17.5 & 0.0 & 0.0 & 76.4 & 35.2 & \textbf{95.6} & 53.8 & 13.8 & 77.5 & 34.3 & 30.2 & 5.0 & 21.8 & 0.0 & 34.4 \\
Biasetton et al. \cite{biasetton2019} & 71.4 & 25.0 & 62.0 & 20.4 & 17.6 & \textbf{26.8} & 5.9 & 0.8 & 64.6 & 24.6 & 86.5 & \textbf{58.3} & \textbf{14.7} & 80.0 & 39.3 & \textbf{42.2} & 5.5 & 22.3 & 0.1 & 35.2 \\\hline
\end{tabular}
\vspace{0.1cm}
\caption{mIoU on the different classes of the Mapillary validation set. The approaches have been trained in a supervised way on the GTA5 dataset and the unsupervised domain adaptation has been performed using the Mapillary training set. \reviewres{The highest value in each column is highlighted in bold.}}
\label{tab:Mapillary_GTA}

\vspace{0.5cm}

\setlength{\tabcolsep}{3.6pt}
\centering
\begin{tabular}{|c|cccccccccccccccc|c|}
\hline
\textbf{SYNTHIA} $\rightarrow$ \textbf{Mapillary}  & \rotatebox{90}{road} &  \rotatebox{90}{sidewalk} &  \rotatebox{90}{building} &  \rotatebox{90}{wall} &\rotatebox{90}{fence} & \rotatebox{90}{pole} & \rotatebox{90}{t light} 
  &\rotatebox{90}{t sign} & \rotatebox{90}{veg} & \rotatebox{90}{sky} & \rotatebox{90}{person}& \rotatebox{90}{rider} & \rotatebox{90}{car} 
 & \rotatebox{90}{bus} &  \rotatebox{90}{mbike} & \rotatebox{90}{bike} & \rotatebox{90}{mean} \\
 \hline
Supervised ($\mathcal{L}_{G,1}$ only) & 14.7 & 18.6 & 34.6 & \textbf{5.4} & \textbf{0.1} & 28.5 & \textbf{0.0} & 0.4 & 73.8 & 62.9 & 50.0 & \textbf{11.4} & \textbf{74.3} & \textbf{28.7} & \textbf{14.0} & 8.1 & 26.6  \\
Ours (full) & \textbf{57.6} & 18.3 & \textbf{62.1} & 0.4 & 0.0 & 23.7 & \textbf{0.0} & 0.0 & \textbf{79.4} & 94.8 & \textbf{52.4} & 9.2 & 74.2 & 28.3 & 4.0 & 6.9 & \textbf{32.0}  \\\hline
Hung et al. \cite{hung2018} & 36.8 & \textbf{20.1} & 53.9 & 0.0 & 0.0 & 23.7 & \textbf{0.0} & 0.0 & 73.9 & \textbf{95.6} & 43.4 & 0.1 & 64.6 & 19.0 & 0.4 & 0.5 & 27.0 \\
Biasetton et al. \cite{biasetton2019} & 16.4 & 19.1 & 42.2 & 2.7 & 0.0 & \textbf{33.1} & \textbf{0.0} & \textbf{1.3} & 76.5 & 88.0 & 50.4 & 10.9 & 69.9 & 25.5 & 6.1 & \textbf{9.2} & 28.2 \\
\hline
\end{tabular}
\vspace{0.1cm}
\caption{mIoU on the different classes of the Mapillary validation set. The approaches have been trained in a supervised way on the SYNTHIA dataset and the unsupervised domain adaptation has been performed using the Mapillary training set. \reviewres{The highest value in each column is highlighted in bold.}}
\label{tab:Mapillary_SYNTHIA}
\end{table*}

\subsection{Evaluation on the Mapillary dataset}

To ensure that our approach can generalize to other real datasets, we performed the same experimental evaluation procedure also on the Mapillary dataset.
We started by using the GTA5 dataset for the supervised training as before. By simply performing a supervised training on GTA5 and then testing on the Mapillary dataset a mIoU of $32.7\%$ can be obtained.
The proposed approach allows to obtain a much more accurate classification with a mIoU of $38.5\%$. Notice that the gain of almost $6\%$ is consistent with the results obtained on the Cityscapes dataset, proving that the performance of the approach \reviewtxt{is} stable across different datasets. The improvement can also be appreciated on both small and large classes, the mIoU values of 14 out of 19 classes show a clear gain. This is also visible in the qualitative results depicted in Figure \ref{fig:qual_res}b, where most of the artifacts on the road surface present in the synthetic trained network disappear and the shape of the small objects is more accurate. The results of \cite{hoffman2016} and \cite{zhang2017} are not available for this dataset, however notice how the approach outperforms by a large margin both \cite{hung2018} and the old version of the approach \cite{biasetton2019} that are  able to reduce only partially the artifacts  on the road surface (visible in all the images), on the cars (row 1) and on the buildings (row 3). 

Furthermore, we can appreciate that also on Mapillary the accuracy is lower \reviewtxt{when adapting from SYNTHIA} leading to a mIoU of $26.6\%$ only. As for Cityscapes, \textit{road} and \textit{sidewalk} classes have an extremely low accuracy due to the poor texture representation (the visual results are reported in the last 3 rows of Figure \ref{fig:qual_res}b). 
By exploiting the proposed unsupervised domain adaptation strategy the mIoU increases to  $32.0\%$ with an improvement of $5.4\%$, again consistent with the other experiments. In this case, the performance is more unstable across the various classes but it is noticeable the large gains on \textit{road} and \textit{building} classes. This is also confirmed by the qualitative results, for example we can appreciate that the proposed approach is the only one able to achieve an accurate and reliable recognition of the \reviewtxt{road.} 
The method of Hung et al. \cite{hung2018} achieves a mIoU of $27\%$ with a very limited improvement w.r.t. the synthetic supervised training. It is strongly penalized by the poor performance on the small and uncommon classes.
The approach of \cite{biasetton2019} has slightly better performance ($28.4\%$), but it has a quite large gap with respect to the proposed method. The weighting scheme and the region growing strategy introduced in this work allowed to obtain a large improvement in this setting.

\subsection{Ablation Study}
\label{sec:ablation}
In this section, we are going to analyze the contributions of the various components \reviewtxt{of the proposed approach.} 
\reviewtxt{We focus on Cityscapes as} target dataset for this study. 
The results of this analysis are shown in Table \ref{tab:ablation_GTA}. By training the generator with a synthetic supervised  approach, i.e., using only $\mathcal{L}_{G,1}$, it is possible to obtain a mIoU of $27.9\%$ when GTA5 is the source dataset and $25.4\%$ when \reviewtxt{adapting from SYNTHIA.} As mentioned in the previous sections, this is the \reviewtxt{less performing  approach.}
A slight improvement can be obtained by adding the adversarial term $\mathcal{L}_{G,2}$ in the loss function. In this case, the mIoU increases of  $1.5\%$ and $2\%$ when the source datasets are GTA5 and SYNTHIA respectively. 
The use of the self-teaching loss $\mathcal{L}_{G,3}$ is particularly useful when exploiting the SYNTHIA dataset, obtaining an improvement of almost $3\%$, probably because the domain shift from this dataset is larger. In the case of GTA5, the improvement is smaller but still significant. 
Moving to the new elements introduced in this work, the region growing strategy (i.e., \reviewtxt{masking with the extended mask $m^R_{T_u}$ instead of using $m_{T_u}$)} allows to a further performance enhancement, especially when using GTA5 with a $2.2\%$ increase, mostly due to the improved handling of medium and large size objects.
When starting from SYNTHIA the gain is more limited but still noticeable (almost $1\%$).
The usage of the discriminator \reviewtxt{output as a weighting factor for the self-teaching loss, without masking with} $m^R_{T_u}$, has a more unstable behavior. This leads to a very good improvement of $2.4\%$ when starting from GTA5 but having almost no impact when employed alone in the SYNTHIA case.
\reviewres{Moreover, we can observe that removing the class weighting term $W^c_t$ in Eq.~\ref{eq:L_G3}, i.e., assigning the same weight to all the classes, we obtain a mIoU of $33.1\%$ and $30.2\%$ when adapting from GTA5 and SYNTHIA, respectively. The values are not too far from the mIoU of the complete version of the approach ($33.3\%$ and $31.3\%$, respectively). This proves that the performances are quite stable with respect to the setting of the weights for the various classes: the approach can work well even if the class frequencies on real data do not accurately match the statistics of synthetic data.}
\reviewtxt{Notice how the complete version of our approach has the best performance. In particular  the  discriminator-based weighting on the SYNTHIA dataset, that alone had a limited impact, is useful when combined with the region growing scheme. }

\reviewres{Finally, Table \ref{tab:ablation_lambdas_GTA} analyzes the impact of the weights that control the relative relevance of the 3 losses. It is possible to notice that the most critical parameter is the weight of the adversarial loss on the source domain $w^s$, that has the largest impact on the final accuracy while the performance are more stable with respect to the other two parameters.}

\newcommand{\dpiVal}{100}

\begin{table}[htbp]
\centering
\setlength{\tabcolsep}{2.6pt}
\begin{tabular}{cccccccc}
$\mathcal{L}_{G,1}$ & $\mathcal{L}_{G,2}$ & $\mathcal{L}_{G,3}$ & \makecell{Region\\Growing} & \makecell{Disc.\\Weight.} & \makecell{Class Weight.\\($W_c^t$)}  & \makecell{mIoU\\GTA} & \makecell{mIoU\\SYNTHIA} \\ \hline
\checkmark   &      &      &     & & & 27.9 & 25.4 \\ \hline
\checkmark   & \checkmark    &      &     & & & 29.4 & 27.4 \\ \hline
\checkmark   &   \checkmark  & \checkmark    &    &    & & 30.4 &  30.2 \\ \hline
\checkmark   &   \checkmark   &  \checkmark    & \checkmark   & & \checkmark & 32.6 & 31.0 \\ \hline
\checkmark   & \checkmark    & \checkmark    &    & \checkmark & \checkmark & 32.8 & 30.2 \\ \hline
\checkmark   & \checkmark & \checkmark &  \checkmark  & \checkmark &  & 33.1 & 30.2 \\ \hline
\checkmark   & \checkmark & \checkmark &  \checkmark  & \checkmark & \checkmark & \textbf{33.3} & \textbf{31.3} \\ \hline
\end{tabular}
\vspace{0.1cm}
\caption{\reviewres{Ablation study on Cityscapes validation set. We analyze the influence of the losses $\mathcal{L}_{G,1}$, $\mathcal{L}_{G,2}$, $\mathcal{L}_{G,3}$, region growing, discriminator weighting and class weighting $W_c^t$.}}
\label{tab:ablation_GTA}
\end{table}

\begin{table}[htbp]
\centering
\setlength{\tabcolsep}{2pt}
\begin{tabular}{|c|c|c|c|c|}
\hline
$w^s$ & $w^t$ & $w'$ & mIoU from GTA5 & mIoU from SYNTHIA \\ \hline \hline 
$\mathbf{\times 1}$ & $\mathbf{\times 1}$ & $\mathbf{\times 1}$ & $\mathbf{33.3}$ & $\mathbf{31.3}$ \\\hline \hline
$\times 1$ & $\times 1$ & $\times 10$ & $32.5$ & $29.8$ \\
$\times 1$ & $\times 1$ & $\times 0.1$ & $32.4$ & $29.2$ \\ \hline \hline

$\times 1$ & $\times 10$ & $\times 1$ & $31.0$ & $28.8$ \\
$\times 1$ & $\times 0.1$ & $\times 1$ & $30.8$ & $24.4$ \\ \hline \hline

$\times 10$ & $\times 1$ & $\times 1$ & $27.2$ & $23.3$ \\
$\times 0.1$ & $\times 1$ & $\times 1$ & $30.8$ & $23.4$ \\ \hline

$\times 4$ & $\times 1$ & $\times 1$ & $29.1$ & $29.0$ \\
$\times 0.25$ & $\times 1$ & $\times 1$ & $32.5$ & $25.2$ \\ \hline

$\times 2$ & $\times 1$ & $\times 1$ & $30.1$ & $30.5$ \\
$\times 0.5$ & $\times 1$ & $\times 1$ & $32.3$ & $28.1$ \\ \hline

\end{tabular}
\vspace{0.1cm}
\caption{\reviewres{Ablation study on the Cityscapes validation set when adapting from GTA5. Different values of the balancing hyper-parameters of Eq.~\ref{eq:L_full} are reported applying various scaling factors to each parameter. The default parameters are $w^s=10^{-2}$, $w^t=10^{-4}$, $w'=10^{-3}$ when adapting from GTA5 and $w^s=10^{-2}$, $w^t=10^{-3}$, $w'=10^{-1}$ when adapting from SYNTHIA.}}
\label{tab:ablation_lambdas_GTA}
\end{table}
\section{Conclusions}
\label{sec:conclusions}

In this paper, \reviewtxt{a novel scheme} to perform unsupervised domain adaptation from synthetic urban scenes to real world ones has been proposed. Two different strategies have been used to exploit unlabeled data. \reviewtxt{The first is based on an adversarial learning framework, based on a fully convolutional discriminator. The second is a soft self-teaching strategy,} based on the assumption that predictions labeled as highly confident by the discriminator are reliable. Additionally, we improved this approach with a region growing module that further refines the confidence maps on the basis of the segmentation output on real-world images.
Experimental results on the  Cityscapes and Mapillary datasets prove the effectiveness of the proposed approach. In particular, we obtained good results both on large sized classes, thanks to the region growing procedure, and on particularly challenging small and uncommon ones, thanks to the class frequency  weighting of the self-teaching loss. 

Further research will be devoted to test the proposed framework with different backbone networks and to the exploitation of generative models to produce more realistic and refined synthetic training data to be fed to the framework.



%

\ifCLASSOPTIONcaptionsoff
  \newpage
\fi



\bibliographystyle{IEEEtran}
\bibliography{strings,refs}
%



%

\begin{IEEEbiography}[{\includegraphics[width=1in,height=1in,clip,keepaspectratio]{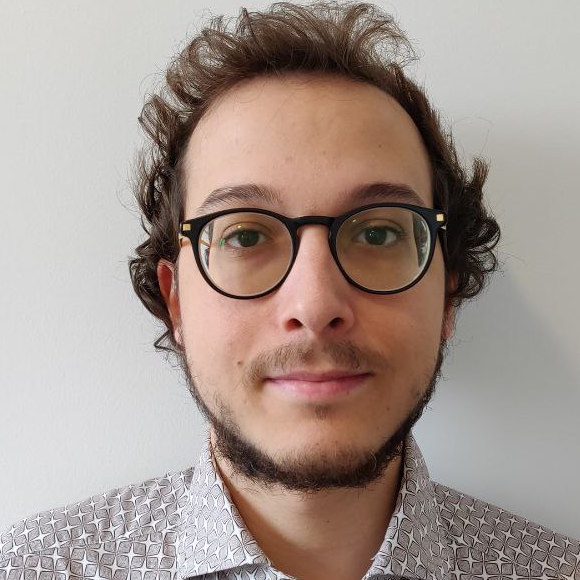}}]{Umberto Michieli}
received the M.Sc. degree in Telecommunication Engineering from the University of Padova in 2018. He is currently a Ph.D. student at the Department of Information Engineering of the University of Padova. In 2018, he spent six months as a Visiting Researcher with the Technische Universit{\"a}t Dresden. His research focuses on transfer learning techniques for semantic segmentation, in particular on domain adaptation and on incremental learning.
\end{IEEEbiography}

\begin{IEEEbiography}[{\includegraphics[width=1in,height=1in,clip,keepaspectratio]{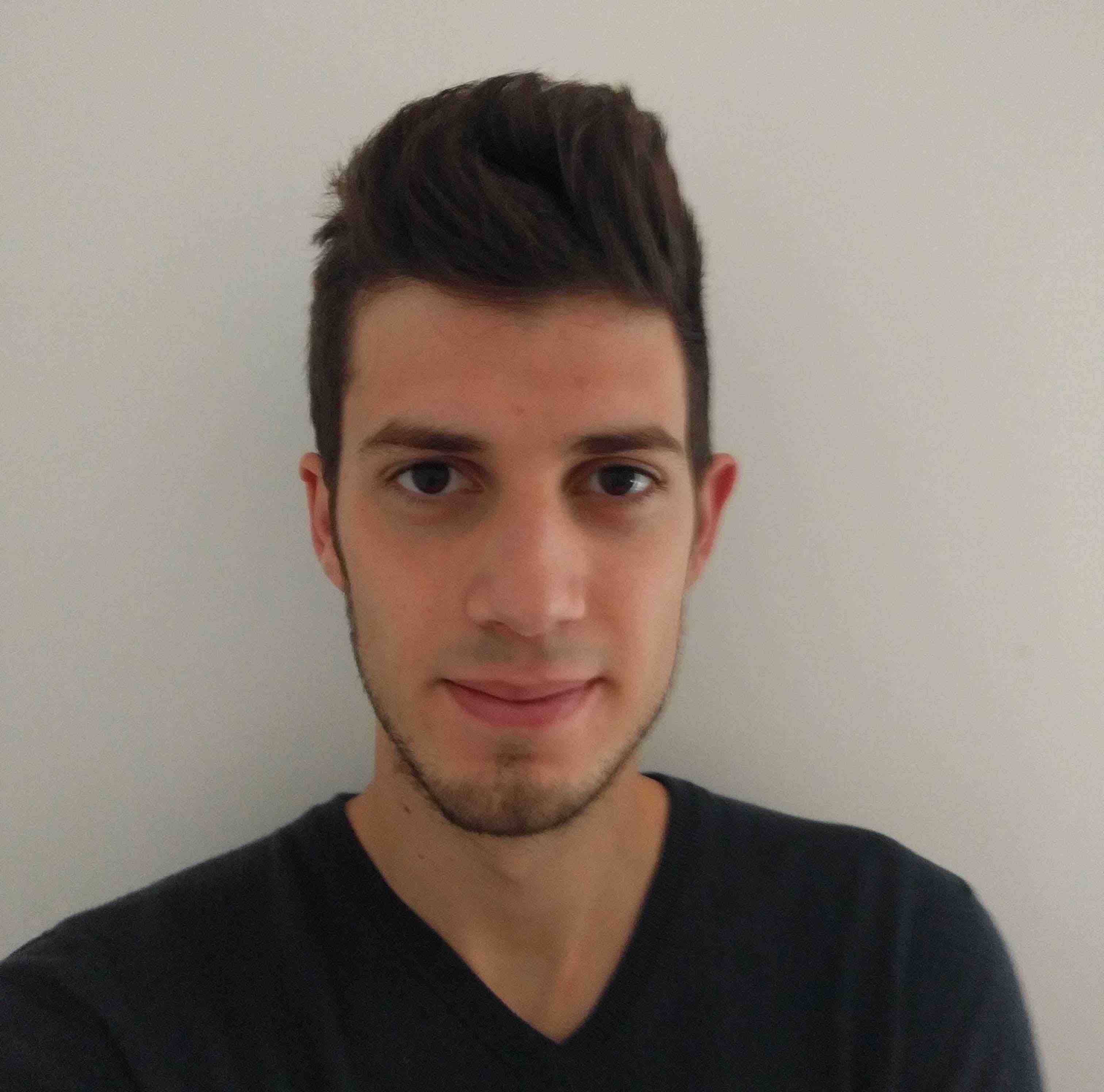}}]{Matteo Biasetton}
received the B.Sc. degree in Information Engineering and the M.Sc. degree in Computer Engineering from the University of Padova, Italy, in 2016 and 2019, respectively. His Master thesis focused on domain adaptation for semantic segmentation. Currently, he is working as a Software Engineer in image processing at Microtec srl.
\end{IEEEbiography}

\begin{IEEEbiography}[{\includegraphics[width=1in,height=1in,clip,keepaspectratio]{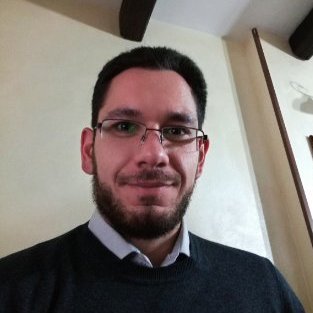}}]{Gianluca Agresti}
received the M.Sc. degree in Telecommunication Engineering from University of Padova in 2016. Currently, he is a PhD candidate at the Department of Information Engineering of University of Padova. His research focuses on deep learning for ToF sensor data processing and multiple sensor fusion for 3D acquisition.
\end{IEEEbiography}

\begin{IEEEbiography}[{\includegraphics[width=1in,height=1in,clip,keepaspectratio]{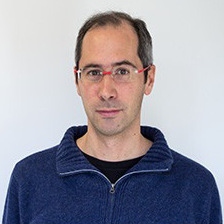}}]{Pietro Zanuttigh}
received a Master degree in Computer Engineering at the University of Padova in 2003 where he also got the Ph.D. degree in 2007. Currently he is an assistant professor at the Department of Information Engineering. His research activity focuses on 3D data processing, in particular ToF sensors data processing, multiple sensors fusion for 3D acquisition, semantic segmentation and hand gesture recognition.
\end{IEEEbiography}







\end{document}